%% file: 0-main.tex
\documentclass[letterpaper, 10 pt, journal, twoside]{IEEEtran} 

\IEEEoverridecommandlockouts                              %

\usepackage[backend=biber,
            url=false,
            isbn=false,
            doi=false,
            backref=false,
            style=ieee,
            natbib=true,%
            mincitenames=1,
            maxcitenames=1,
            citestyle=numeric-comp,
            sorting=nyt,%
            block=none]{biblatex}
        
\renewcommand{\bibfont}{\small}
\input{0-preamble.tex}
\input{0-defs.tex}

\usepackage{cleveref}
\title{
Safety Augmented Value Estimation from Demonstrations (SAVED): Safe Deep Model-Based RL for Sparse Cost Robotic Tasks
}

\author{Brijen Thananjeyan$^{*1}$, Ashwin Balakrishna$^{*1}$, Ugo Rosolia$^{2}$, Felix Li$^{1}$, Rowan McAllister$^{1}$, \\ Joseph E. Gonzalez$^{1}$, Sergey Levine$^{1}$, Francesco Borrelli$^{1}$, Ken Goldberg$^{1}$
\thanks{$^{*}$ Equal contribution}%
\thanks{Manuscript received: September, 10, 2019; Revised January, 2, 2020; Accepted February, 7, 2020.}%
\thanks{This paper was recommended for publication by Editor Dongheui Lee upon evaluation of the Associate Editor and Reviewers' comments.}
\thanks{$^{1}$Brijen Thananjeyan, Ashwin Balakrishna, Felix Li, Rowan McAllister, Joseph E. Gonzalez, Sergey Levine, Francesco Borrelli, and Ken Goldberg are with the Dept. of Electrical Engineering and Computer Science, University of California, Berkeley, USA
        {\tt\footnotesize \{bthananjeyan, ashwin\_balakrishna, fzli, rmcallister, jegonzal, slevine, fborrelli, goldberg\}@berkeley.edu}}%
\thanks{$^{2} $Ugo Rosolia is with the Dept. of Mechanical and Civil Engineering, California Institute of Technology, USA
        {\tt\footnotesize urosolia@caltech.edu}}
\thanks{Digital Object Identifier (DOI): see top of this page.}
}

\begin{document}

\maketitle

\begin{abstract}

Reinforcement learning (RL) for robotics is challenging due to the difficulty in hand-engineering a dense cost function, which can lead to unintended behavior, and dynamical uncertainty, which makes exploration and constraint satisfaction challenging. We address these issues with a new model-based reinforcement learning algorithm, \algname, which uses supervision that only identifies task completion and a modest set of suboptimal demonstrations to constrain exploration and learn efficiently while handling complex constraints. We then compare \algabbr with $3$ state-of-the-art model-based and model-free RL algorithms on $6$ standard simulation benchmarks involving navigation and manipulation and a physical knot-tying task on the da Vinci surgical robot. Results suggest that \algabbr outperforms prior methods in terms of success rate, constraint satisfaction, and sample efficiency, making it feasible to safely learn a control policy directly on a real robot in less than an hour. For tasks on the robot, baselines succeed less than $5 \%$ of the time while \algabbr has a success rate of over $75\%$ in the first 50 training iterations. Code and supplementary material is available at \url{https://tinyurl.com/saved-rl}.

\end{abstract}


\input{1-introduction.tex}
\input{2-related-work.tex}
\input{3-assumptions_prelims.tex}

\input{4-SAVED.tex}

\input{5-method.tex}

\input{7-experiments.tex}
\input{8-discussion.tex}

\input{9-acknowledgements.tex}

\renewcommand*{\bibfont}{\footnotesize}
\printbibliography %
\clearpage
\input{9.5-supp-material-arxiv.tex}

\end{document}

%% file: 0-preamble.tex
\usepackage{graphics}
\usepackage{dsfont}
\usepackage[pdftex]{graphicx}
\usepackage{wrapfig}
\DeclareGraphicsExtensions{.pdf,.png,.jpg}
\usepackage{epsfig}
\usepackage[font={small}]{caption}
\usepackage{subcaption}
\usepackage[rightcaption]{sidecap}
\usepackage{pbox}

\usepackage{bigstrut}
\usepackage{mathtools}
\usepackage{amsmath, amssymb, amscd}
\usepackage{ wasysym } %
\usepackage{mathptmx} %
\usepackage{gensymb} 
\usepackage{nicefrac}       %
\numberwithin{equation}{section} 

\DeclareMathAlphabet{\mathcal}{OMS}{lmsy}{m}{n}
\usepackage{textcomp} %

\usepackage{algorithm} %
\usepackage{algorithmicx, algpseudocode} %

\usepackage{array} %
\usepackage{tabularx}
\usepackage{multirow}
\usepackage{multicol}
\usepackage{booktabs}
\usepackage{tabulary}

\usepackage[utf8]{inputenc}
\usepackage[english]{babel} %
\usepackage{units}
\usepackage{bm}
\usepackage{xspace}
\usepackage{flushend}%
\usepackage{balance} 
\usepackage{csquotes}
\usepackage{makeidx}
\usepackage{blindtext}

\usepackage{enumitem}

\usepackage{ragged2e}
\usepackage{soul} %
\usepackage{subfiles} %

\usepackage{url}
\makeatletter
\g@addto@macro{\UrlBreaks}{\UrlOrds}
\makeatother
\usepackage{color}
\usepackage[usenames,dvipsnames,table,xcdraw]{xcolor}
\usepackage{pgfplots} %
\pgfplotsset{compat=newest}

\usepackage{marginnote}
\usepackage{soul} %

\usepackage[colorinlistoftodos]{todonotes}
\newcommand{\tocite}[1]{%
\textcolor{red}{[cite:\ifthenelse{\equal{#1}{}}{}{#1}?]}
}

\newcommand{\ignore}[1]{}

\renewcommand{\baselinestretch}{1.0}



\usepackage[pdfborder={0 0 0.5}]{hyperref}
\hypersetup{
    colorlinks=true,
    linkcolor=black,
    citecolor=black,
    filecolor=cyan,
    urlcolor=black
}

%% file: 0-defs.tex
\newcommand{\myargmin}[1] {\underset{#1}{\text{argmin }}}

\newcommand{\algname}{Safety Augmented Value Estimation from Demonstrations (SAVED)\xspace}
\newcommand{\algnamecap}{Safety Augmented Value Estimation from Demonstrations (SAVED)\xspace}

\newcommand{\algabbr}{SAVED\xspace}
\renewcommand{\theequation}{\thesubsection.\arabic{equation}}

%% file: 1-introduction.tex
\section{Introduction}
To use RL in the real world, algorithms need to be efficient, easy to use, and safe, motivating methods which are reliable even with significant dynamical uncertainty. Deep model-based reinforcement learning (deep MBRL) is of significant interest because of its sample efficiency advantages over model-free methods in a variety of tasks, such as assembly, locomotion, and manipulation~\cite{NNDynamics, handful-of-trials, MPCRacing, PILCO, Lenz2015DeepMPCLD, converging-supervisor, OneShotMBRL}. However, past work in deep MBRL typically requires dense hand-engineered cost functions, which are hard to design and can lead to unintended behavior~\cite{amodei2016concrete}. It would be easier to simply specify task completion in the cost function, but this setting is challenging due to the lack of expressive supervision. This motivates using demonstrations, which allow the user to roughly specify desired behavior without extensive engineering effort. However, providing high-performing trajectories of the task may be challenging, motivating methods that can rapidly improve upon suboptimal demonstrations that may be supplied via a PID controller or kinesthetically. Furthermore, in many robotic tasks, specifically in domains such as surgery, safe exploration is critical to ensure that the robot does not damage itself or cause harm to its surroundings. To enable this, deep MBRL algorithms must be able to satisfy user-specified (and possibly nonconvex) state-space constraints.

\begin{figure}
\centering
\begin{subfigure}[t]{0.47\textwidth}
  \centering
  \includegraphics[width=\linewidth]{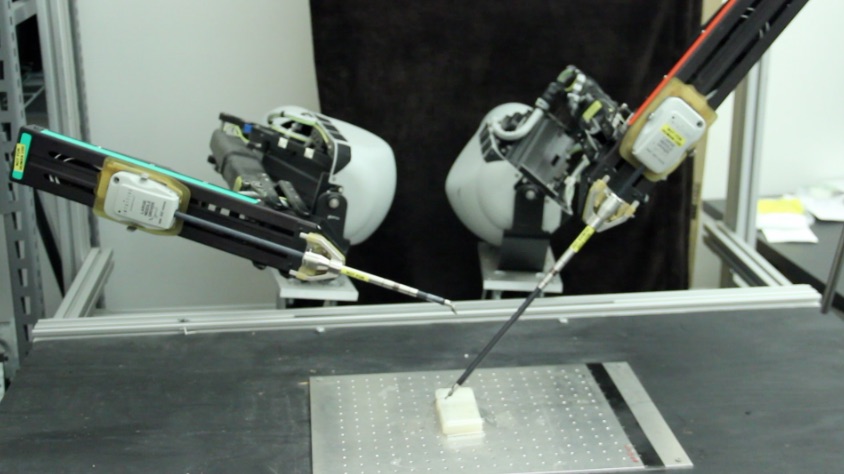}
  \label{dvrk}
\end{subfigure}
\vfill
\begin{subfigure}[t]{0.47\textwidth}
  \centering
  \includegraphics[width=\linewidth]{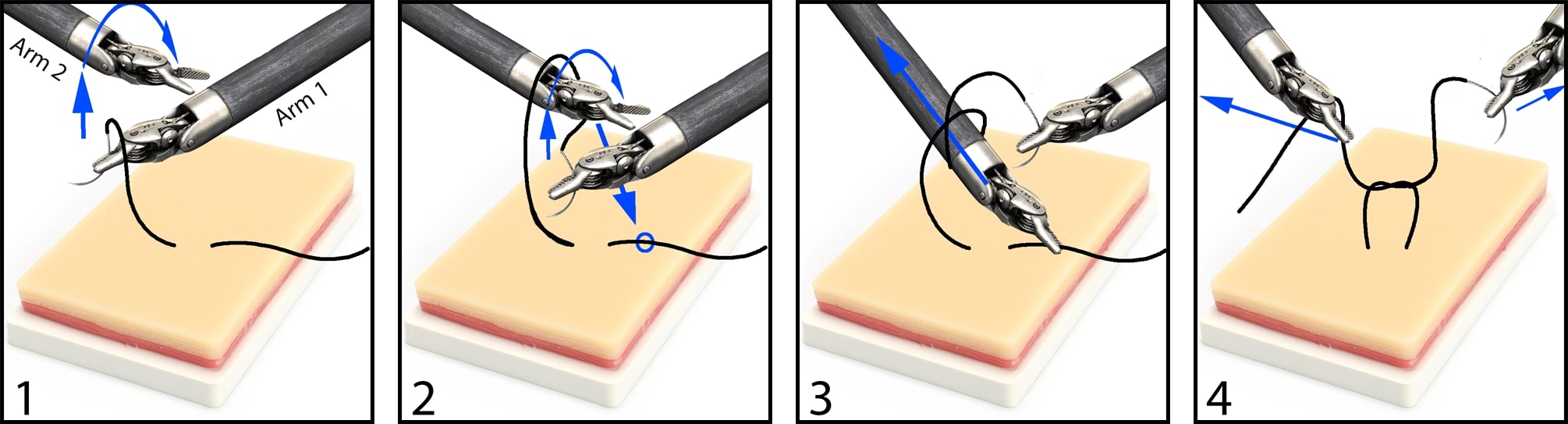}
  \label{knot-tying-dvrk}
\end{subfigure}
\caption{\algabbr is able to safely learn maneuvers on the da Vinci surgical robot, which is difficult to precisely control~\cite{cable-driven}. We demonstrate that \algabbr is able to optimize inefficient human demonstrations of a surgical knot-tying task, substantially improving on demonstration performance with just 15 training iterations.}
\label{SAVED-figure}
\end{figure}

We develop a method to efficiently use deep MBRL in dynamically uncertain environments with both sparse costs and complex constraints. We address the difficulty of hand-engineering cost functions by using a small number of suboptimal demonstrations to provide a signal about task progress in sparse cost environments, which is updated based on agent experience. Then, to enable stable policy improvement and constraint satisfaction, we impose two probabilistic constraints to (1) constrain exploration by ensuring that the agent can plan back to regions in which it is confident in task completion and (2) leverage uncertainty estimates in the learned dynamics to implement chance constraints~\cite{nemirovski2012safe} during learning. The probabilistic implementation of constraints makes this approach broadly applicable, since it can handle settings with significant dynamical uncertainty, where enforcing constraints exactly is difficult.

We introduce a new algorithm motivated by deep model predictive control (MPC) and robust control, \algname, which enables efficient learning for sparse cost tasks given a small number of suboptimal demonstrations while satisfying the provided constraints. We specifically consider tasks with a tight start state distribution and fixed, known goal set. \algabbr is evaluated on MDPs with unknown dynamics, which are iteratively estimated from experience, and with a cost function that only indicates task completion. The contributions of this work are (1) a novel method for constrained exploration driven by confidence in task completion, (2) a technique for leveraging model uncertainty to probabilistically enforce complex constraints, enabling obstacle avoidance or optimizing demonstration trajectories while maintaining desired properties, (3) experimental evaluation against $3$ state-of-the-art model-free and model-based RL baselines on $8$ different environments, including simulated experiments and  physical maneuvers on the da Vinci surgical robot. Results suggest that \algabbr achieves superior sample efficiency, success rate, and constraint satisfaction rate across all domains considered and can be applied efficiently and safely for learning directly on a real robot. 

%% file: 2-related-work.tex
\section{Related work}
There is significant interest in model-based planning and deep MBRL \cite{PILCO, Lenz2015DeepMPCLD, OneShotMBRL, Plan-Online-Learn-Offline, handful-of-trials, NNDynamics} due to the improvements in sample efficiency when planning over learned dynamics compared to model-free methods for continuous control \cite{SAC, TD3}. However, most prior deep MBRL algorithms use hand-engineered dense cost functions to guide exploration and planning, which we avoid by using demonstrations to provide signal about delayed costs. Demonstrations have been leveraged to accelerate learning for a variety of model-free RL algorithms, such as Deep Q Learning \cite{LFD-DQN} and DDPG \cite{LFD-DDPG, overcoming-exp}, but model-free methods are typically less sample efficient and cannot anticipate constraint violations since they learn reactive policies~\cite{GapMBRLMFRL}. Demonstrations have also been leveraged in model-based algorithms, such as in motion planning with known dynamics~\cite{elastic-bands} and for seeding a learned dynamics model for fast online adaptation using iLQR and a dense cost~\cite{OneShotMBRL}, distinct from the task completion based costs we consider. Unlike traditional motion planning algorithms, which generate open-loop plans to a goal configuration when dynamics are known, we consider designing a closed-loop controller that operates in stochastic dynamical systems where the system dynamics are initially unknown and iteratively estimated from data. Finally,~\citet{TREX} use inverse RL to significantly outperform suboptimal demonstrations, but do not enforce constraints or consistent task completion during learning. 

In iterative learning control (ILC), the controller tracks a predefined reference trajectory and data from each iteration is used to improve closed-loop performance~\cite{bristow2006survey}. \citet{StochasticMPC, SampleBasedLMPC, LearningMPC} provide a reference-free algorithm to iteratively improve the performance of an initial trajectory by using a safe set and terminal cost to ensure recursive feasibility, stability, and local optimality given a known, deterministic nonlinear system or stochastic linear system under certain regularity assumptions. In contrast to \citet{StochasticMPC, SampleBasedLMPC, LearningMPC}, we consider designing a similar controller in stochastic non-linear dynamical systems where the dynamics are unknown and iteratively estimated from experience. Thus, \algabbr uses function approximation to estimate a dynamics model, value function, and safe set. There has also been significant interest in safe RL \cite{SafeRLSurvey}, typically focusing on exploration while satisfying a set of explicit constraints~\cite{SafeRLWithSupervision, SafeExpInMDPs, CPO}, satisfying specific stability criteria \cite{StabilitySafeMBRL}, or formulating planning via a risk sensitive Markov Decision Process \cite{RiskAversion, RobustRiskSensitiveMDP}. Distinct from prior work in safe RL and control, \algabbr can be successfully applied in settings with both uncertain dynamics and sparse costs by using probabilistic constraints to constrain exploration to feasible regions during learning.

%% file: 3-assumptions_prelims.tex
\section{\algnamecap}
\label{SAVED-summary}
This section describes how \algabbr uses a set of suboptimal demonstrations to constrain exploration while satisfying user-specified state space constraints. First, we discuss how \algabbr learns system dynamics and a value function to guide learning in sparse cost environments. Then, we motivate and discuss the method used to enforce constraints under uncertainty to both ensure task completion during learning and satisfy user-specified state space constraints. 
\subsection{Assumptions and Preliminaries}
\label{assumptions-preliminaries}
In this work, we consider stochastic, unknown dynamical systems with a cost function that only identifies task completion. We outline the framework for MBRL using a standard Markov Decision Process formulation. A finite-horizon Markov Decision Process (MDP) is a tuple $(\mathcal{X}, \mathcal{U}, P(\cdot, \cdot), T, C(\cdot,\cdot))$ where $\mathcal{X}$ is the feasible (constraint-satisfying) state space and $\mathcal{U}$ is the control space. The stochastic dynamics model $P$ maps a state and control input to a probability distribution over states, $T$ is the task horizon, and $C$ is the cost function. A stochastic control policy $\pi$ maps an input state to a distribution over $\mathcal{U}$.

We assume that (1) tasks are iterative in nature, and have a fixed low-variance start state distribution and fixed, known goal set $\mathcal{G}$. This is common in a variety of repetitive tasks, such as assembly, surgical knot-tying, and suturing. We further assume that (2) the user specifies an indicator function $\mathds{1}\left(x \in \mathcal{X}\right)$, which checks whether a state $x$ is constraint-satisfying. Finally, we assume that (3) a modest set of suboptimal but constraint satisfying demos are available, for example from imprecise human teleoperation or a hand-tuned PID controller. This enables rough specification of desired behavior without having to design a dense cost function, allowing us to consider cost functions which only identify task completion: $C(x, u) = \mathds{1}_{\mathcal{G}^C}(x)$, where $\mathcal{G} \subset \mathcal{X}$ defines a goal set in the state space and $\mathcal{G}^C$ is its complement. We define task success by convergence to $\mathcal{G}$ at the end of the task horizon without violating constraints. Note that under this definition of costs, the problem we consider is equivalent to the shortest time control problem in optimal control, but with initially unknown system dynamics which are iteratively estimated from experience. The applicability of \algabbr extends beyond this particular choice of cost function, but we focus on this class due to its convenience and notorious difficulty for reinforcement learning algorithms~\cite{overcoming-exp}.

Finally, we define the notion of a safe set to enable constrained policy improvement, which is described further in Section \ref{safety-estimation-methods}. Recent MPC literature~\cite{LearningMPC} motivates constraining exploration to regions in which the agent is confident in task completion, which gives rise to desirable theoretical properties when dynamics are known and satisfy certain regularity conditions~\cite{StochasticMPC, SampleBasedLMPC, LearningMPC}. For a trajectory at iteration $k$, given by $x^k = \{x^k_t | t \in \mathbb{N}\}$, we define the \textit{sampled safe set} as 
\begin{small}
\begin{align}
\label{eq:safeSet}
\mathcal{SS}^j = \bigcup_{k \in \mathcal{M}^j} x^k
\end{align}
\end{small}
where $\mathcal{M}^j = \{k\in[0,j):\lim_{t\rightarrow\infty}x_t^k \in \mathcal{G}\}$ is the set of indices of all successful trajectories before iteration $j$ as in \citet{LearningMPC}. $\mathcal{SS}^j$ contains the states from all iterations before $j$ from which the agent controlled the system to $\mathcal{G}$ and is initialized from demonstrations. A key operating principle of \algabbr is to use $\mathcal{SS}^j$ to guide exploration by ensuring that there always exists a way to plan back into a continuous approximation of $\mathcal{SS}^j$. This allows for policy improvement while ensuring that the agent can always return to a state from which it has previously completed the task, enabling consistent task completion during learning.

%% file: 4-SAVED.tex
\begin{figure*}
\centering
\includegraphics[width=0.99\textwidth]{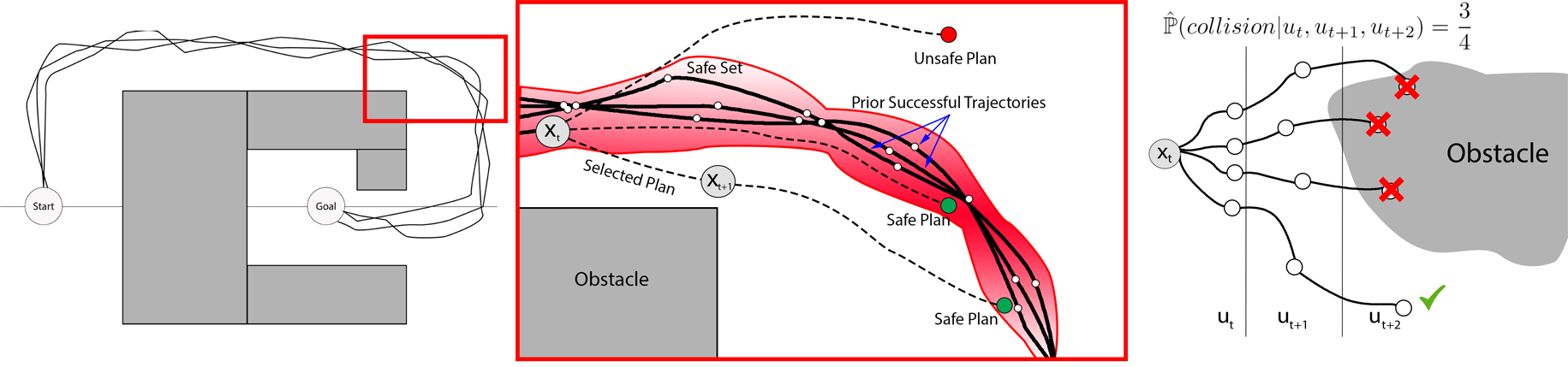}
\caption{\textbf{Task Completion Driven Exploration (left):} A density model is used to represent the region in state space where the agent has high confidence in task completion; trajectory samples over the learned dynamics that do not have sufficient density at the end of the planning horizon are discarded. The agent may explore outside the safe set as long as a plan exists to guide the agent back to the safe set from the current state; \textbf{Chance Constraint Enforcement (right): } Implemented by sampling imagined rollouts over the learned dynamics for the same sequence of controls multiple times and estimating the probability of constraint violation by the percentage of rollouts that violate a constraint.}
\label{SAVED-methods}
\end{figure*}

\subsection{Algorithm Overview}
\subsubsection{Deep Model Predictive Control}
\algabbr uses MPC to optimize costs over a sequence of controls at each state. However, when using MPC, since the current control is computed by solving a finite-horizon approximation to the infinite-horizon control problem, an agent may take shortsighted controls which may make it impossible to complete the task safely, such as planning the trajectory of a race car over a short horizon without considering an upcoming curve~\cite{borrelli2017predictive}. Additionally, the planner receives no feedback or information about task progress when using the indicator task functions used in this work. Thus, to guide exploration in temporally-extended tasks, we solve the problem in equation~\ref{eq-objective-1}, which includes a learned value function in the objective. Note that $\mathcal{U}^H$ refers to the set of $H$ length control sequences while $\mathcal{X}^{H+1}$ refers to the set of $H+1$ length state sequences. This corresponds to the standard objective in MPC with an appended value function $V^\pi_\phi$, which provides a terminal cost estimate for the current policy at the end of the planning horizon. 

While prior work in deep MBRL \cite{NNDynamics, handful-of-trials} has primarily focused on planning over learned dynamics, we introduce a learned value function, which is initialized from demonstrations to provide initial signal, to guide planning even in sparse cost settings. The learned dynamics model $f_\theta$ and value function $V_\phi^\pi$ are each represented with a finite probabilistic ensemble of $n$ neural networks (in this work we pick $n = 5$), as is used to represent system dynamics in \citet{handful-of-trials}. The probabilistic ensemble consists of a set of neural networks, each of which output the parameters of a conditional axis-aligned Gaussian distribution and are trained on bootstrapped samples from the training dataset using a maximum likelihood objective as in~\cite{handful-of-trials}. Each conditional Gaussian is used to model aleatoric uncertainty in the dynamics, while the bootstrapped ensemble of these models captures epistemic uncertainty due to data availability in different regions of the MDP. \algabbr uses the learned stochasticity of the models to enforce probabilistic constraints when planning under uncertainty. These functions are initialized from demonstrations and updated from data collected from each training iteration. See supplementary material for further details on how these networks are trained.

\subsubsection{Probabilistic Constraints}
The core novelties of \algabbr are the additional probabilistic constraints in \ref{eq-objective-3} to encourage task completion driven exploration and enforce user-specified chance constraints. First, a non-parametric density model $\rho$ is trained on $\mathcal{SS}^j$, which includes states from prior successful trajectories, including those from demonstrations. $\rho$ constraints exploration by requiring $x_{t+H}$ to fall in a region with high probability of task completion. This enforces cost-driven constrained exploration and iterative improvement, enabling reliable performance even with sparse costs. Note that the agent can still explore new regions, as long as it has a plan that can take it back to the safe set with high probability. Second, we require all elements of $x_{t:t+H}$ to fall in the feasible region $\mathcal{X}^{H+1}$ with probability at least $\beta$, which enables probabilistic enforcement of state space constraints. In Section \ref{safety-estimation-methods}, we discuss the methods used for task completion driven exploration and in Section \ref{constraint-enforcement-methods}, we discuss how probabilistic constraints are enforced during learning.

In summary, \algabbr solves the following optimization problem at each timestep based on the current state of the system, $x_t$, which is measured from observations:

\begin{small}
\begin{subequations}
\label{eq-objective-full}
\begin{align}
u_{t:t+H-1}^* &= \myargmin{u_{t:t+H-1}\in \mathcal{U}^H} \mathds{E}_{x_{t:t+H}}\left[\sum_{i=0}^{H -1} C(x_{t+i}, u_{t+i}) + V_\phi^\pi (x_{t+H})\right] \label{eq-objective-1}\\
\text{s.t. } & x_{t+i+1} \sim f_\theta(x_{t+i}, u_{t+i})\ \forall i \in \{0,\ldots,H-1\} \label{eq-objective-2}\\
& \rho_{\alpha}(x_{t+H}) > \delta, \mathds{P}\left(x_{t:t+H} \in \mathcal{X}^{H+1}\right) \geq \beta \label{eq-objective-3}
\end{align}
\end{subequations}
\end{small}

\begin{algorithm}[htb!]
  \caption{\algnamecap}
  \label{alg:main}
\begin{algorithmic}
  \Statex{\textbf{Require:} Replay Buffer $\mathcal{R}$; value function $V^\pi_\phi(x)$, dynamics model $\hat{f_{\theta}}(x' | x, u)$, and safety density model $\rho_\alpha(x)$ all seeded with demos; kernel and chance constraint parameters $\alpha$ and $\beta$.
}
  \For{$i \in \{1,\dots,N\}$}
    \State{Sample $x_0$ from start state distribution}
      \For{$t \in \{1,\dots,T-1\}$}
          \State Pick $u^{*}_{t:t+H-1}$ by solving eq. \ref{eq-objective-full} using CEM
          \State Execute $u^*_t$ and observe $x_{t+1}$
          \State $\mathcal{R} = \mathcal{R} \cup \{(x_t, u^*_t, C(x_{t}, u^*_{t}), x_{t+1})\}$
      \EndFor
      \If{$x_T \in \mathcal{G}$}
            \State Update safety density model $\rho_{\alpha}$ with $x_{0 : T}$
        \EndIf
      \State{Optimize $\theta$ and $\phi$ with $\mathcal{R}$}
    \EndFor
\end{algorithmic}
\end{algorithm}

%% file: 5-method.tex
\subsection{Task Completion Driven Exploration}
\label{safety-estimation-methods}
Under certain regularity assumptions, if states at the end of the MPC planning horizon are constrained to fall in the sampled safe set $\mathcal{SS}^j$, iterative improvement, controller feasibility, and convergence are guaranteed given known stochastic linear dynamics or deterministic nonlinear dynamics ~\cite{LearningMPC, StochasticMPC, SampleBasedLMPC}. The way we constrain exploration in \algabbr builds on this prior work, but we note that unlike \citet{LearningMPC, StochasticMPC, SampleBasedLMPC}, \algabbr is designed for settings in which dynamics are completely unknown, nonlinear, and stochastic. As illustrated in Figure~\ref{SAVED-methods}, the mechanism for constraining exploration allows the agent to generate trajectories that leave the safe set as long as a plan exists to navigate back in, enabling policy improvement. By adding newly successful trajectories to the safe set, the agent is able to further improve its performance. Note that since the safety density model and value function are updated on-policy, the support of the safety density model expands over iterations, while the value function is updated to reflect the current policy. This enables \algabbr to improve upon the performance of the demonstrations since on each iteration, it simply needs to be able to plan back to the high support region of a safety density model fit on states from which \algabbr was able to complete the task from \textit{all prior iterations} rather than just those visited by the demonstrations.

Since $\mathcal{SS}^j$ is a discrete set, we introduce a continuous approximation by fitting a density model $\rho$ to $\mathcal{SS}^j$. Instead of requiring that $x_{t+H} \in \mathcal{SS}^j$, \algabbr instead enforces that $\rho_{\alpha}(x_{t+H}) > \delta$, where $\alpha$ is a kernel width parameter (constraint \ref{eq-objective-3}). Since the tasks considered in this work have sufficiently low ($< 17$) state space dimension, kernel density estimation provides a reasonable approximation. We implement a top-hat kernel density model using a nearest neighbors classifier with a tuned kernel width $\alpha$ and use $\delta = 0$ for all experiments. Thus, all states within Euclidean distance $\alpha$ from the closest state in $\mathcal{SS}^j$ are considered safe under $\rho_{\alpha}$, representing states in which the agent is confident in task completion. As the policy improves, it may forget how to complete the task from very old states in $\mathcal{SS}^j$, so such states are evicted from $\mathcal{SS}^j$ to reflect the current policy when fitting $\rho_{\alpha}$. We discuss how these constraints are implemented in Section \ref{constraint-enforcement-methods}, with further details in the supplementary material. In future work, we will investigate implicit density estimation techniques to scale to high-dimensional settings.

\subsection{Chance Constraint Enforcement}
\label{constraint-enforcement-methods}
\algabbr leverages uncertainty estimates in the learned dynamics to enforce probabilistic constraints on its trajectories. This allows \algabbr to handle complex, user-specified state space constraints to avoid obstacles or maintain certain properties of demonstrations without a user-shaped or time-varying cost function. During MPC trajectory optimization, control sequences are sampled from a truncated Gaussian distribution that is iteratively updated using the cross-entropy method (CEM)~\cite{handful-of-trials}. Each control sequence is simulated multiple times over the stochastic dynamics model as in~\cite{handful-of-trials} and the average return of the simulations is used to score the sequence. However, unlike \citet{handful-of-trials}, we implement chance constraints by discarding control sequences if more than $100 \cdot (1 - \beta)\%$ of the simulations violate constraints (constraint \ref{eq-objective-3}), where $\beta$ is a user-specified tolerance. Note that the $\beta$ parameter controls the tradeoff between ensuring sufficient exploration to learn the dynamics and satisfying specified constraints. This is illustrated in Figure \ref{SAVED-methods}. The task completion constraint (Section \ref{safety-estimation-methods}) is implemented similarly, with control sequences discarded if any of the simulated rollouts do not terminate in a state with sufficient density under $\rho_{\alpha}$. 

\subsection{Algorithm Pseudocode}
We summarize \algabbr in Algorithm \ref{alg:main}. The dynamics, value function, and state density model are initialized from suboptimal demonstrations. At each iteration, we sample a start state and then controls are generated by solving equation \ref{eq-objective-full} using the cross-entropy method (CEM) at each timestep. Transitions are collected in a replay buffer to update the dynamics, value function, and safety density model at the end of each iteration. The state density model is only updated if the last trajectory was successful.

%% file: 7-experiments.tex
\section{Experiments}
\label{experiments}

We evaluate \algabbr on simulated continuous control benchmarks and on real robotic tasks with the da Vinci Research Kit (dVRK) \cite{kazanzides-chen-etal-icra-2014} against state-of-the-art deep RL algorithms and find that \algabbr outperforms all baselines in terms of sample efficiency, success rate, and constraint satisfaction rate during learning. All tasks use $C(x, u) = \mathds{1}_{\mathcal{G}^C}(x)$ (Section \ref{assumptions-preliminaries}), which yields a controller which maximizes the time spent inside the goal set. All algorithms are given the same demonstrations and are evaluated by measuring iteration cost, success rate, and constraint satisfaction rate (if applicable). Tasks are only considered successfully completed if the agent reaches and stays in $\mathcal{G}$ until the end of the episode. Constraint violation results in early termination of the episode.

For all experiments, we run each algorithm 3 times to control for stochasticity in training and plot the mean iteration cost vs. time with error bars indicating the standard deviation over the 3 runs. Additionally, when reporting task success rate and constraint satisfaction rate, we show bar plots indicating the median value over the 3 runs with error bars between the lowest and highest value over the 3 runs. When reporting the iteration cost of \algabbr and all baselines, any constraint violating trajectory is reported by assigning it the maximum possible iteration cost $T$, where $T$ is the task horizon. Thus, any constraint violation is treated as a catastrophic failure. We plan to explore soft constraints as well in future work. Furthermore, for all simulated tasks, we also report best achieved iteration costs, success rates, and constraint satisfaction rates for model-free methods after 10,000 iterations since they take much longer to start performing the task even when supplied with demonstrations. 

For \algabbr, dynamics models and value functions are each represented with a probabilistic ensemble of 5, 3 layer neural networks with 500 hidden units per layer with swish activations as used in \citet{handful-of-trials}. To plan over the dynamics, the TS-$\infty$ trajectory sampling method from \cite{handful-of-trials} is used. We use 5 and 30 training epochs for dynamics and value function training when initializing from demonstrations. When updating the models after each training iteration, 5 and 15 epochs are used for the dynamics and value functions respectively. Value function initialization is done by training the value function using the true cost-to-go estimates from demonstrations. However, when updated on-policy, the value function is trained using temporal difference error (TD-1) on a replay buffer containing prior states. The safety density model, $\rho$, is trained on a fixed history of states from which the agent was able to reach the goal (safe states), where this history can be tuned based on the experiment (see supplement). We represent the density model using kernel density estimation with a tophat kernel. Instead of modifying $\delta$ for each environment, we set $\delta= 0$ (keeping points with positive density), and modify $\alpha$ (the kernel parameter/width), which works well in practice. See the supplementary material for additional experiments, videos, and ablations with respect to choice of $\alpha$, $\beta$, and demonstration quantity/quality. We also include further details on baselines, network architectures, hyperparameters, and training procedures.

\subsection{Baselines}
We consider the following set of model-free and model-based baseline algorithms. To enforce constraints for model-based baselines, we augment the algorithms with the simulation based method described in Section \ref{constraint-enforcement-methods}. Because model-free baselines have no such mechanism to readily enforce constraints, we instead apply a very large cost when constraints are violated. See supplementary material for an ablation of the reward function used for model-free baselines.

\begin{enumerate}[topsep=0pt,
noitemsep]
    \item \textbf{Behavior Cloning (Clone)}: Supervised learning on demonstration trajectories.
    \item \textbf{PETS from Demonstrations (PETSfD)}: Probabilistic ensemble trajectory sampling (PETS) from  Chua et al~\cite{handful-of-trials} with the dynamics model initialized with demo trajectories and planning horizon long enough to plan to the goal (judged by best performance of \algabbr).
    \item \textbf{PETSfD Dense}: PETSfD with tuned dense cost.
    \item \textbf{Soft Actor Critic from Demonstrations (SACfD)}: Model-free RL algorithm, Soft Actor Critic~\cite{SAC}, where demo transitions are used for training initially.
    \item \textbf{Overcoming Exploration in Reinforcement Learning from Demonstrations (OEFD)}: Model-free algorithm from \citet{overcoming-exp} which combines model-free RL with a behavior cloning loss on the demonstrations to accelerate learning.
    \item \textbf{\algabbr (No SS)}: \algabbr without the \textit{sampled safe set} constraint described in Section \ref{safety-estimation-methods}.
\end{enumerate}

\begin{figure*}[!htb]
\vspace{10pt}
\begin{subfigure}[t]{.99\textwidth}
  \centering
  \includegraphics[width=\linewidth]{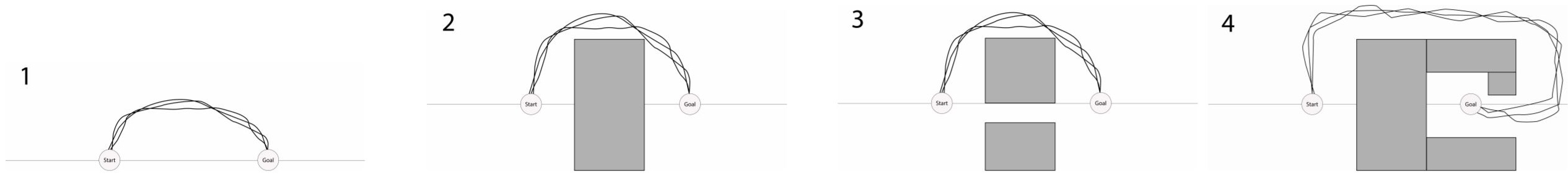}
\end{subfigure}%
\vfill
\begin{subfigure}[t]{.99\textwidth}
  \centering
  \includegraphics[width=\linewidth]{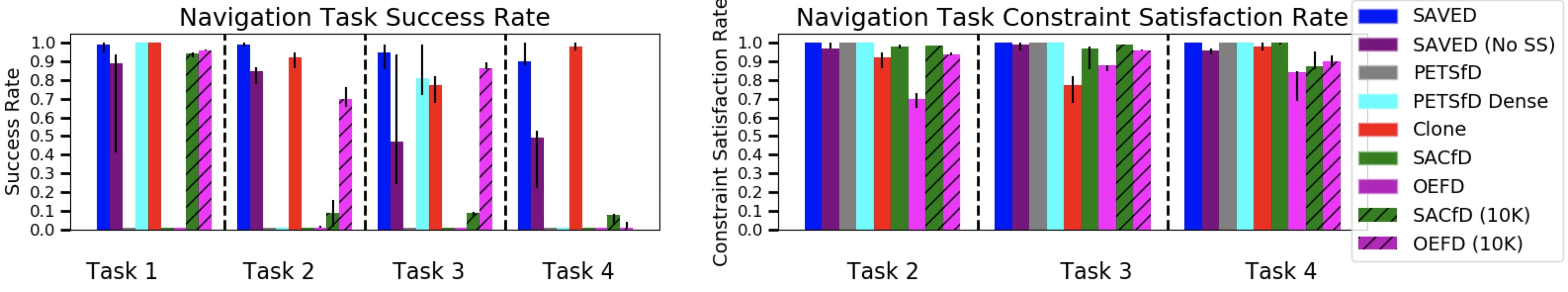}
\end{subfigure}%
\vfill
\begin{subfigure}[t]{.99\textwidth}
  \centering
  \includegraphics[width=\linewidth]{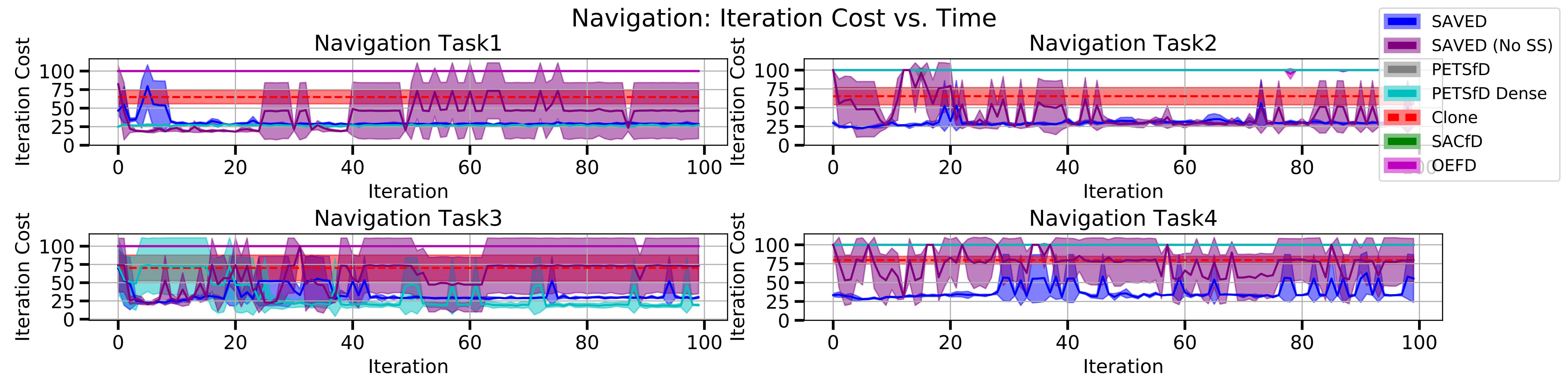}
\end{subfigure}%

\caption{\textbf{Navigation Domains:} \algabbr is evaluated on 4 navigation tasks. Tasks 2-4 contain obstacles, and task 3 contains a channel for passage to $\mathcal{G}$ near the x-axis. \algabbr learns significantly faster than all RL baselines on tasks 2 and 4. In tasks 1 and 3, \algabbr has lower iteration cost than baselines using sparse costs, but does worse than PETSfD Dense, which is given dense Euclidean norm costs to find the shortest path to the goal. For each task and algorithm, we report success and constraint satisfaction rates over the first 100 training iterations and also over the first 10,000 iterations for SACfD and OEFD. We observe that \algabbr has higher success and constraint satisfaction rates than other RL algorithms using sparse costs across all tasks, and even achieves higher rates in the first 100 training iterations than model-free algorithms over the first 10,000 iterations.}
\label{pointbotExperiments}
\end{figure*}

\subsection{Simulated Navigation}

To evaluate whether \algabbr can efficiently and safely learn temporally extended tasks with nonconvex constraints, we consider a 4-dimensional ($x$, $y$, $v_x$, $v_y$) navigation task in which a point mass is navigating to a goal set, which is a unit ball centered at the origin. The agent can exert force in cardinal directions and experiences drag coefficient $\psi$ and Gaussian process noise $z_t\sim \mathcal{N}(0, \sigma^2 I)$ in the dynamics. 
We use $\psi = 0.2$ and $\sigma = 0.05$ in all experiments in this domain. Demonstrations trajectories are generated by guiding the robot along a very suboptimal hand-tuned trajectory for the first half of the trajectory before running LQR on a quadratic approximation of the true cost. Gaussian noise is added to the demonstrator policy. Additionally, we use a planning horizon of 15 for \algabbr and 25, 30, 30, 35 for PETSfD for tasks 1-4 respectively.
\begin{figure*}[!htb]
\centering
\includegraphics[width=0.99\linewidth]{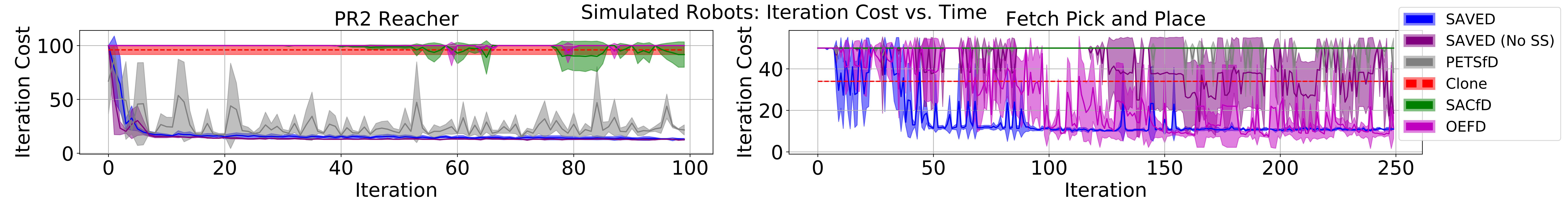}
\caption{\textbf{Simulated Robot Experiments Performance: }\algabbr achieves better performance than all baselines on both tasks. We use 20 demonstrations with average iteration cost of 94.6 for the reacher task and 100 demonstrations with average iteration cost of 34.4 for the pick and place task. For the reacher task, the safe set constraint does not improve performance, likely because the task is very simple, but for pick and place, we see that the safe set constraint adds significant training stability.}
\label{simulatedRobotExperiments}
\end{figure*}
The 4 experiments run on this environment are:
\begin{enumerate}[topsep=0pt,
noitemsep]
\item \textbf{Long navigation task to the origin: }$x_0 = (-100, 0)$ We use 50 demonstrations with average return of $73.9$ and kernel width $\alpha = 3$.
\item \textbf{Large obstacle blocking the $x$-axis: }This environment is difficult for approaches that use a Euclidean norm cost function due to local minima. We use 50 demonstrations with average return of $67.9$, kernel width $\alpha = 3$, and chance constraint parameter $\beta = 1$. 
\item \textbf{Large obstacle with a small channel near the $x$-axis: }This environment is difficult for the algorithm to optimally solve since the iterative improvement of paths taken by the agent is constrained. We use $x_0 = (-50, 0)$, 50 demonstrations with average return of $67.9$, kernel width $\alpha = 3$, and chance constraint parameter $\beta = 1$. 
\item \textbf{Large obstacle surrounds the goal set with a small channel for entry: } This environment is very difficult to solve without demonstrations. We use $x_0 = (-50, 0)$, $100$ demonstrations with average return of $78.3$, kernel width $\alpha = 3$, and chance constraint parameter $\beta = 1$.
\end{enumerate}

\algabbr has a higher success rate than all other RL baselines using sparse costs, even including model-free baselines over the first 10,000 iterations, while never violating constraints across all navigation tasks. Furthermore, this performance advantage is amplified with task difficulty. Only Clone and PETSfD Dense ever achieve a higher success rate, but Clone does not improve upon demonstration performance (Figure \ref{pointbotExperiments}) and PETSfD Dense has additional information about the task. Furthermore, \algabbr learns significantly more efficiently than all RL baselines on all navigation tasks except for tasks 1 and 3, in which PETSfD Dense with a Euclidean norm cost function finds a better solution. While \algabbr (No SS) can complete the tasks, it has a much lower success rate than \algabbr, especially in environments with obstacles as expected, demonstrating the importance of the \textit{sampled safe set} constraint. Note that SACfD, OEFD, and PETSfD make essentially no progress in the first 100 iterations and never complete any of the tasks in this time, although they mostly satisfy constraints. After 10,000 iterations of training, SACfD and OEFD achieve average best iteration costs of $23.7$ and $23.8$ respectively on task 1, $21$ and $21.7$ respectively on task 2, $17.3$ and $19$ respectively on task 3, and $23.7$ and $40$ respectively on task 4. Thus, we see that \algabbr achieves comparable performance in the first 100 iterations to the asymptotic performance of model-free RL algorithms while maintaining consistent task completion and constraint satisfaction during learning.

\begin{figure*}
\centering
\begin{subfigure}[t]{.31\textwidth}
  \centering
  \includegraphics[width=\linewidth]{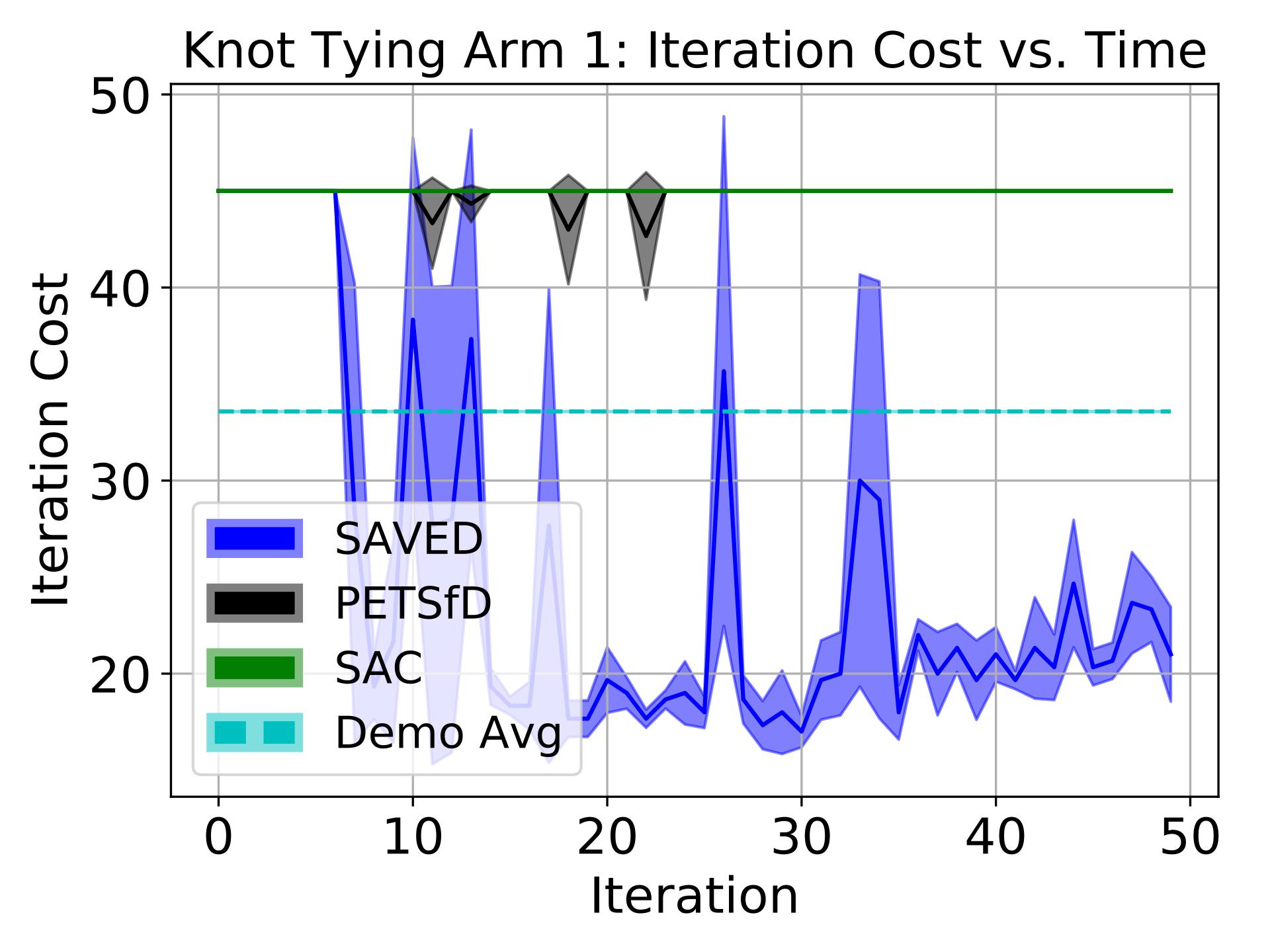}
\end{subfigure}
\hfill
\begin{subfigure}[t]{.31\textwidth}
  \centering
  \includegraphics[width=1.1\linewidth]{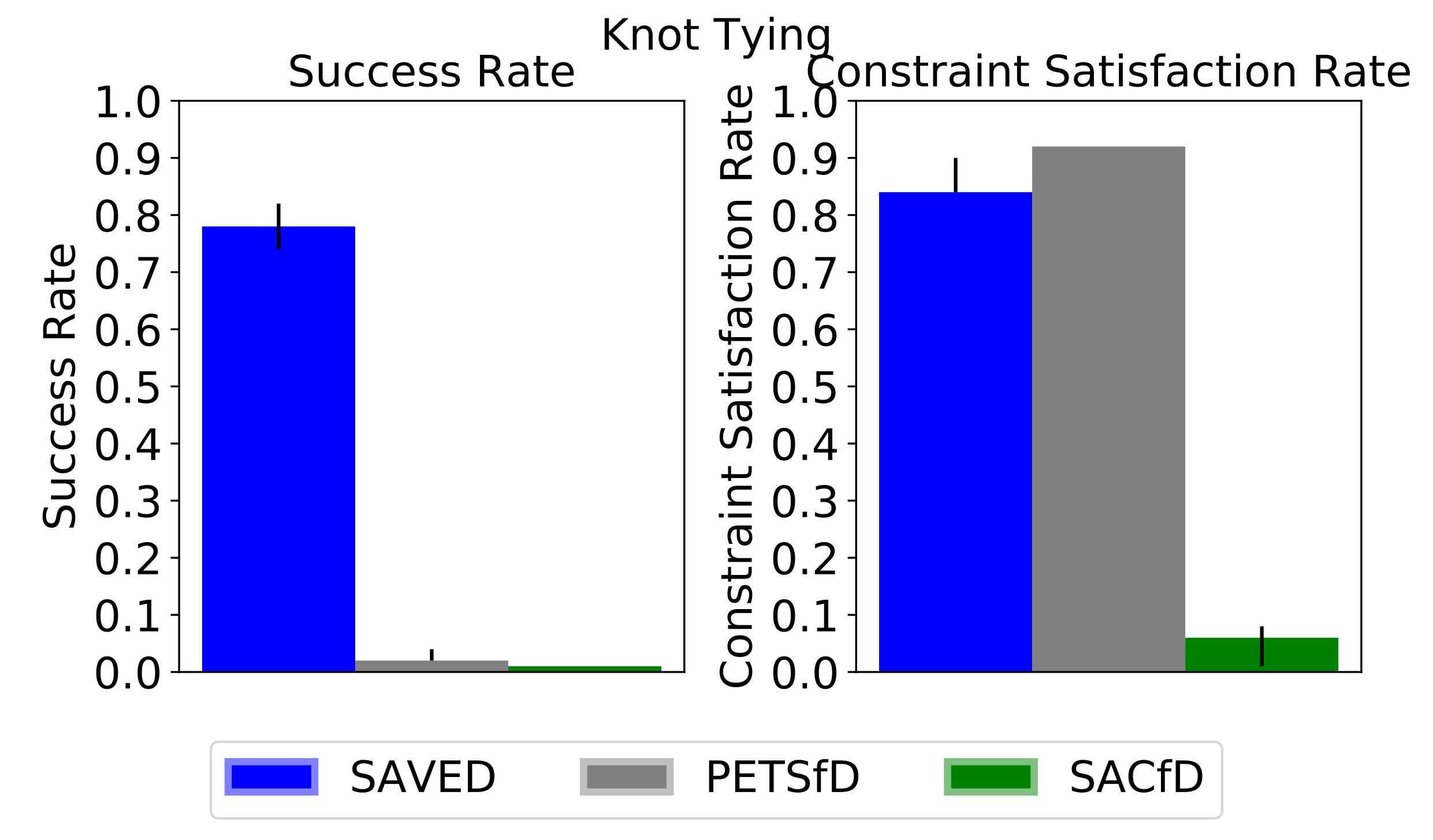}
\end{subfigure}
\hfill
\begin{subfigure}[t]{.31\textwidth}
  \centering
  \includegraphics[width=\linewidth]{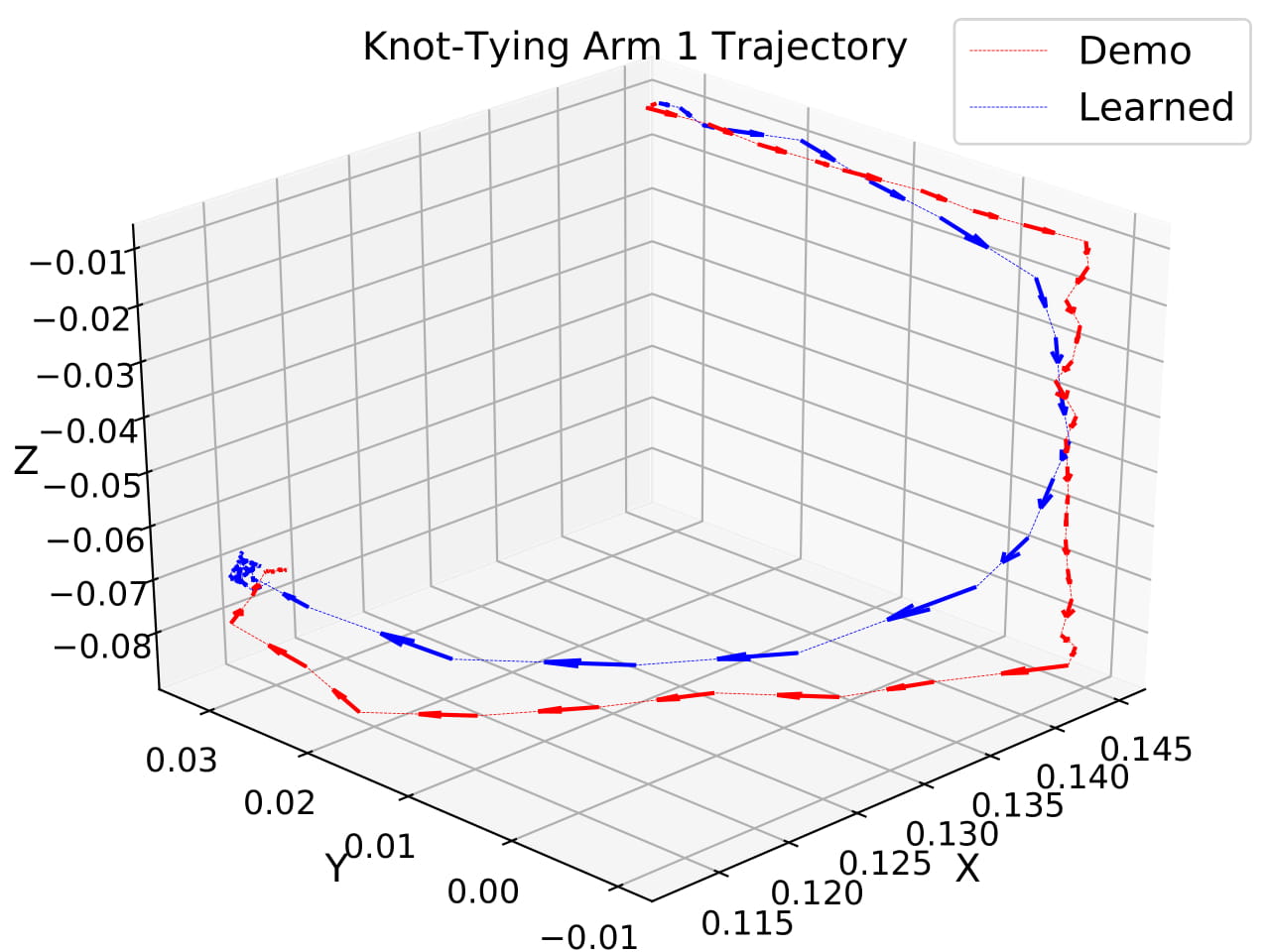}
  \label{knot-tying-trajs}
\end{subfigure}%
\caption{
\textbf{Physical Surgical Knot-Tying: }\textbf{Training Performance: }After just $15$ iterations, the agent completes the task relatively consistently with only a few failures, and converges to a iteration cost of $22$, faster than demos, which have an average iteration cost of $34$. In the first $50$ iterations, both baselines mostly fail, and are less efficient than demos when they do succeed; \textbf{ Trajectories: }\algabbr quickly learns to speed up with only occasional constraint violations.}
\label{knot-tying-plots}
\end{figure*}

\subsection{Simulated Robot Experiments}
To evaluate whether \algabbr also outperforms baselines on standard unconstrained environments, we consider sparse versions of two common simulated robot tasks: the torque-controlled PR2 Reacher environment from \citet{handful-of-trials} with a fixed goal and on a pick and place task with a simulated, position-controlled Fetch robot from \cite{gymrobotics}. The reacher task involves controlling the end-effector of a simulated PR2 robot to a small ball in $\mathds{R}^3$. The state representation consists of $7$ joint positions, $7$ joint velocities, and the goal position. The goal set is specified by a $0.05$m radius Euclidean ball in state space. Suboptimal demonstrations are generated with average cost $94.6$ by training PETS with a shaped cost function that heavily penalizes large torques. We use $\alpha = 15$ for \algabbr and a planning horizon of $25$ for both \algabbr and PETSfD. SACfD and OEFD achieve a best iteration cost of $9$ and $60$ respectively over 10,000 iterations of training averaged over the 3 runs. The pick and place task involves picking up a block from a fixed location on a table and also guiding it to a small ball in $\mathds{R}^3$. The task is simplified by automating the gripper motion, which is difficult for \algabbr to learn due to the bimodality of gripper controls, which is hard to capture with the unimodal truncated Gaussian distribution used during CEM sampling. The state representation for the task consists of (end effector relative position to object, object relative position to goal, gripper jaw positions). Suboptimal demonstrations are generated by hand-tuning a controller that slowly but successfully completes the task with average iteration cost $34.4$. We use a safe set buffer size of $5000$ and $\alpha = 0.05$. We use a planning horizon of $10$ for \algabbr and $20$ for PETSfD. SACfD and OEFD both achieve a best iteration cost of $6$ over 10,000 iterations of training averaged over the 3 runs.

\algabbr learns faster than all baselines on both tasks (Figure \ref{simulatedRobotExperiments}) and exhibits significantly more stable learning in the first 100 and 250 iterations for the reacher and pick and place tasks respectively. However, while \algabbr is substantially more sample efficient than SACfD and OEFD for these tasks, both algorithms achieve superior asymptotic performance.

\subsection{Physical Robot Experiments}

We evaluate the ability of \algabbr to learn a surgical knot-tying task with nonconvex state space constraints on the da Vinci Research Kit (dVRK)~\cite{kazanzides-chen-etal-icra-2014}. The dVRK is cable-driven and has relatively imprecise controls, motivating model learning~\cite{cable-driven}. Furthermore, safety is paramount due to the cost and delicate structure of the arms. The goal of these tasks is to speed up demonstration trajectories while still maintaining properties of the trajectories that result in a task completion. This is accomplished by constraining learned trajectories to fall within a tight, 1 cm tube of the demos. The goal set is represented with a 1 cm ball in $\mathds{R}^3$ and the robot is controlled via delta-position control, with a maximum control magnitude of 1 cm during learning for safety. Robot experiments are very time consuming due to interactive data collection, so training RL algorithms on limited physical hardware is difficult without sample efficient algorithms. We include additional experiments on a Figure-8 tracking task in the supplementary material.

\subsubsection{Surgical Knot-Tying} \algabbr is used to optimize demonstrations of a surgical knot-tying task on the dVRK, using the same multilateral motion as in~\cite{knot-tying-surgery}. Demonstrations are hand-engineered for the task, and then policies are optimized for one arm (arm 1), while a hand-engineered policy is used for the other arm (arm 2). While arm 1 wraps the thread around arm 2, arm 2 simply moves down, grasps the other end of the thread, and pulls it out of the phantom as shown in Figure \ref{SAVED-figure}. Thus, we only expect significant performance gain by optimizing the policy for the portion of the arm 1 trajectory which involves wrapping the thread around arm 2. We only model the motion of the end-effectors in 3D space. We use $\beta = 0.8$, $\alpha = 0.05$, planning horizon $10$, and 100 demonstrations with average cost $34.4$ for \algabbr. We use a planning horizon of $20$ and $\beta=1.$ for PETSfD. \algabbr quickly learns to smooth out demo trajectories, with a success rate of over $75 \%$ (Figure \ref{knot-tying-plots}) during training, while baselines are unable to make sufficient progress in this time. PETSfD rarely violates constraints, but also almost never succeeds, while SACfD almost always violates constraints and never completes the task. Training \algabbr directly on the real robot for $50$ iterations takes only about an hour, making it practical to train on a real robot for tasks where data collection is expensive. At execution-time (post-training), we find that \algabbr is very consistent, successfully tying a knot in $20/20$ trials with average iteration cost of $21.9$ and maximum iteration cost of $25$ for the arm 1 learned policy, significantly more efficient than demos which have an average iteration cost of $34$. See supplementary material for trajectory plots of the full knot-tying trajectory and the Figure-8 task.

%% file: 8-discussion.tex
\section{Discussion and Future Work}
We present \algabbr, a model-based RL algorithm that can efficiently learn a variety of robotic control tasks in the presence of dynamical uncertainty, sparse cost feedback, and complex constraints by using suboptimal demonstrations to constrain exploration to regions in which the agent is confident in task completion. We then empirically evaluate \algabbr on 6 simulated benchmarks and on a knot-tying task on a real surgical robot. Results suggest that \algabbr is more sample efficient and has higher success and constraint satisfaction rates than all RL baselines and can be efficiently and safely trained on a real robot. In future work, we will explore convergence and safety guarantees for \algabbr and extensions to a wide distribution of start states and goal sets. Additionally, a limitation of \algabbr is that solving the MPC objective with CEM makes high frequency control difficult. In future work, we will explore distilling the learned controller into a reactive policy to enable fast policy evaluation in practice.

%% file: 9-acknowledgements.tex
\section{Acknowledgments}
\footnotesize
This research was performed at the AUTOLAB at UC Berkeley in affiliation with the Berkeley AI Research (BAIR) Lab, Berkeley Deep Drive (BDD), the Real-Time Intelligent Secure Execution (RISE) Lab, and the CITRIS "People and Robots" (CPAR) Initiative. Authors were also supported by the Scalable Collaborative Human-Robot Learning (SCHooL) Project, a NSF National Robotics Initiative Award 1734633, and in part by donations from Google and Toyota Research Institute. Ashwin Balakrishna is supported by an NSF GRFP and Ugo Rosolia was partially supported by the Office of Naval Research (N00014-311). This article solely reflects the opinions and conclusions of its authors and do not reflect the views of the sponsors or their associated entities. We thank our colleagues who provided helpful feedback and suggestions, in particular Suraj Nair, Jeffrey Ichnowski, Anshul Ramachandran, Daniel Seita, Marius Wiggert, and Ajay Tanwani.

%% file: 9.5-supp-material-arxiv.tex
\onecolumn
\begin{LARGE}
\begin{center}
\textbf{\algname: Safe Deep Model-Based RL for Sparse Cost Robotic Tasks Supplementary Material}
\end{center}
\end{LARGE}
\section{Additional Experimental Details for \algabbr and Baselines}
For all experiments, we run each algorithm 3 times to control for stochasticity in training and plot the mean iteration cost vs. time with error bars indicating the standard deviation over the 3 runs. Additionally, when reporting task success rate and constraint satisfaction rate, we show bar plots indicating the median value over the 3 runs with error bars between the lowest and highest value over the 3 runs. Experiments are run on an Nvidia DGX-1 and on a desktop running Ubuntu 16.04 with a 3.60 GHz Intel Core i7-6850K, 12 core CPU and an NVIDIA GeForce GTX 1080. When reporting the iteration cost of \algabbr and all baselines, any constraint violating trajectory is reported by assigning it the maximum possible iteration cost $T$, where $T$ is the task horizon. Thus, any constraint violation is treated as a catastrophic failure. We plan to explore soft constraints as well in future work.

\subsection{\algabbr}

\subsubsection{Dynamics and Value Function}
For all environments, dynamics models and value functions are each represented with a probabilistic ensemble of 5, 3 layer neural networks with 500 hidden units per layer with swish activations as used in \citet{handful-of-trials}. To plan over the dynamics, the TS-$\infty$ trajectory sampling method from \cite{handful-of-trials} is used. We use 5 and 30 training epochs for dynamics and value function training when initializing from demonstrations. When updating the models after each training iteration, 5 and 15 epochs are used for the dynamics and value functions respectively. All models are trained using the Adam optimizer with learning rate 0.00075 and 0.001 for the dynamics and value functions respectively. Value function initialization is done by training the value function using the true cost-to-go estimates from demonstrations. However, when updated on-policy, the value function is trained using temporal difference error (TD-1) on a buffer containing all prior states. Since we use a probabilistic ensemble of neural networks to represent dynamics models and value functions, we built off of the provided implementation \cite{PETS_github} of PETS in \cite{handful-of-trials}.

\subsubsection{Constrained Exploration}

Define states from which the system was successfully stabilized to the goal in the past as safe states. We train density model $\rho$ on a fixed history of safe states, where this history is tuned based on the experiment. We have found that simply training on all prior safe states works well in practice on all experiments in this work. We represent the density model using kernel density estimation with a top-hat kernel. Instead of modifying $\delta$ for each environment, we set $\delta= 0$ (keeping points with positive density), and modify $\alpha$ (the kernel parameter/width). We find that this works well in practice, and allows us to speed up execution by using a nearest neighbors algorithm implementation from scikit-learn. We are experimenting with locality sensitive hashing, implicit density estimation as in~\citet{EX2}, and have had some success with Gaussian kernels as well (at significant additional computational cost). The exploration strategy used by \algabbr in navigation task 2 is illustrated in Figure \ref{traj-evolution}.

\subsection{Behavior Cloning}
We represent the behavior cloning policy with a neural network with 3 layers of 200 hidden units each for navigation tasks and pick and place, and 2 layers of 20 hidden units each for the PR2 Reacher task. We train on the same demonstrations provided to \algabbr and other baselines for 50 epochs.
\subsection{PETSfD and PETSfD Dense}
PETSfD and PETSfD Dense use the same network architectures and training procedure as \algabbr and the same parameters for each task unless otherwise noted, but just omit the value function and density model $\rho$ for enforcing constrained exploration. PETSfD uses a planning horizon that is long enough to complete the task, while PETSfD Dense uses the same planning horizon as \algabbr.
\subsection{SACfD}
We use the rlkit implementation \cite{SAC_github} of soft actor critic with the following parameters: batch size=128, discount=$0.99$, soft target $\tau=0.001$, policy learning rate = $3e-4$, Q function learning rate = $3e-4$, and value function learning rate = $3e-4$, batch size = $128$, replay buffer size = $1000000$, discount factor = $0.99$. All networks are two-layer multi-layer perceptrons (MLPs) with 300 hidden units. On the first training iteration, only transitions from demonstrations are used to train the critic. After this, SACfD is trained via rollouts from the actor network as usual. We use a similar reward function to that of \algabbr, with a reward of -1 if the agent is not in the goal set and 0 if the agent is in the goal set. Additionally, for environments with constraints, we impose a reward of -100 when constraints are violated to encourage constraint satisfaction. The choice of collision reward is ablated in section~\ref{mf-ablation}. This reward is set to prioritize constraint satisfaction over task success, which is consistent with the selection of $\beta$ in the model-based algorithms considered.
\subsection{OEFD}
We use the implementation of OEFD provided by \citet{OEFD_github} with the following parameters: learning rate = $0.001$, polyak averaging coefficient = $0.8$, and L2 regularization coefficient = $1$. During training, the random action selection rate is $0.2$ and the noise added to policy actions is distributed as $\mathcal{N}(0,\,1)$. All networks are three-layer MLPs with 256 hidden units. Hindsight experience replay uses the ``future'' goal replay and selection strategy with $k = 4$ \cite{HER}. Here $k$ controls the ratio of HER data to data coming from normal experience replay in the replay buffer. We use a similar reward function to that of \algabbr, with a reward of -1 if the agent is not in the goal set and 0 if the agent is in the goal set. Additionally, for environments with constraints, we impose a reward of -100 when constraints are violated to encourage constraint satisfaction. The choice of collision reward is ablated in section~\ref{mf-ablation}. This reward is set to prioritize constraint satisfaction over task success, which is consistent with the selection of $\beta$ in the model-based algorithms considered.

\section{Simulated Experiments Additional Results}

\begin{figure}[!htb]
  \centering
  \includegraphics[width=0.4\linewidth]{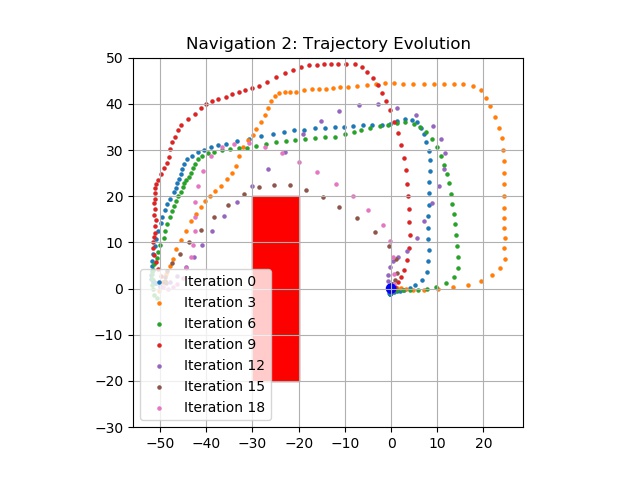}
  \caption{\textbf{Navigation Task 2 Trajectory Evolution: }\algabbr rapidly improves upon demonstration trajectories by constraining its exploration to regions of relative certainty and high cost. By iteration 15, \algabbr is able to find a safe but efficient trajectory to the goal at the origin.}
  \label{traj-evolution}
\end{figure}

In Figure~\ref{traj-evolution}, we illustrate the mechanism by which \algabbr iteratively improves upon suboptimal demonstrations on navigation task 2 by planning into an expanding safe set.

\begin{figure}[!htb]
  \centering
  \includegraphics[width=0.4\linewidth]{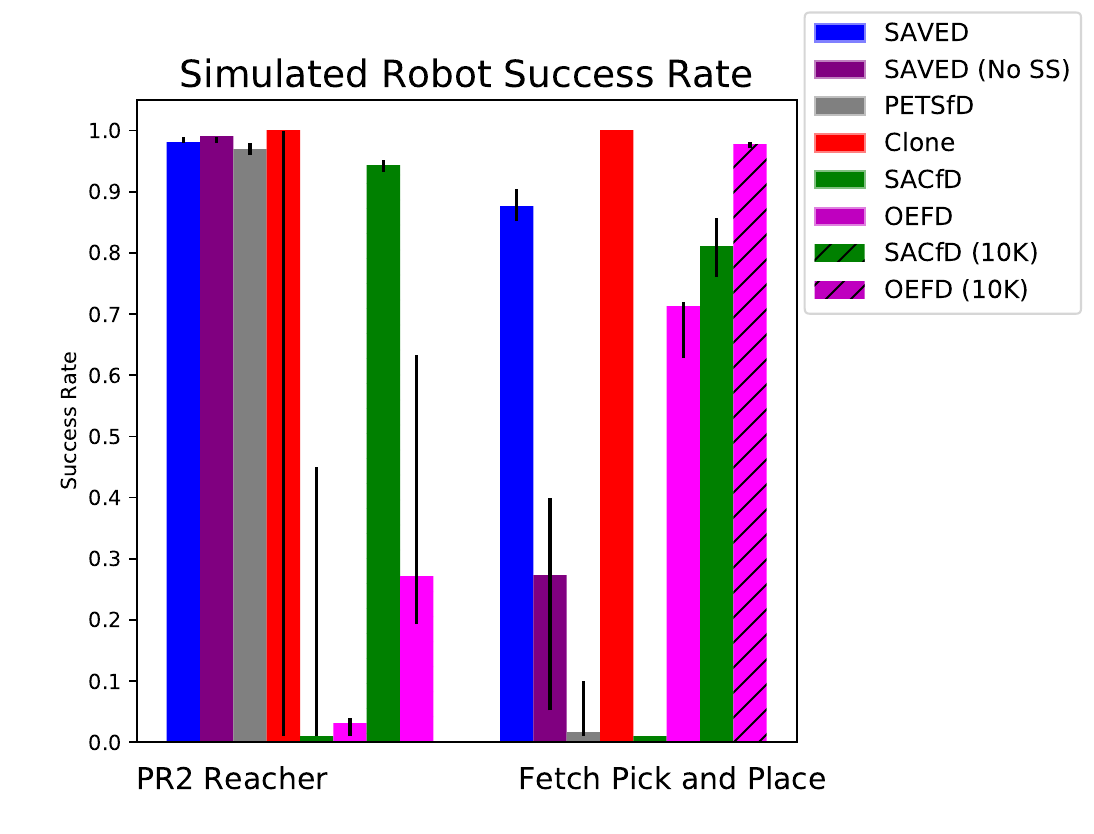}
  \caption{\textbf{Simulated Robot Experiments Success Rate: }\algabbr has comparable success rate to Clone, PETSfD, and \algabbr(No SS) on the reacher task in the first 100 iterations. For the pick and place task, \algabbr outperforms all baselines in the first 250 iterations except for Clone, which does not improve upon demonstration performance.}
  \label{sim-robot-stats}
\end{figure}

In Figure \ref{sim-robot-stats}, we show the task success rate for the PR2 reacher and Fetch pick and place tasks for \algabbr and baselines. We note that \algabbr outperforms RL baselines (except \algabbr (No SS) for the reacher task, most likely because the task is relatively simple so the \textit{sampled safe set} constraint has little effect) in the first 100 and 250 iterations for the reacher and pick and place tasks respectively. Note that although behavior cloning has a higher success rate, it does not improve upon demonstration performance. However, although \algabbr's success rate is not as different from the baselines in these environments as those with constraints, this result shows that \algabbr can be used effectively in a general purpose way, and still learns more efficiently than baselines in unconstrained environments as seen in the main paper.

\section{Physical Experiments: Additional Details and Experiments}
For all experiments, $\alpha = 0.05$ and a set of $100$ hand-coded trajectories with a small amount of Gaussian noise added to the controls is generated. For all physical experiments, we use $\beta = 1$ for PETSfD since we found this gave the best performance when no signal from the value function was provided. In all tasks, the goal set is represented with a 1 cm ball in $\mathbb{R}^3$. The dVRK is controlled via delta-position control, with a maximum control magnitude set to 1 cm during learning for safety. We train state density estimator $\rho$ on all prior successful trajectories for the physical robot experiments. 

\subsection{Figure-8}
In addition to the knot-tying task discussed in the main paper, we also evaluate \algabbr on a Figure-8 tracking task on the surgical robot. In this task, the dVRK must track a Figure 8 in the workspace. The agent is constrained to remain within a 1 cm pipe around a reference trajectory with chance constraint parameter $\beta = 0.8$ for \algabbr and $\beta = 1$ for PETSfD. We use $100$ inefficient but successful and constraint-satisfying demonstrations with average iteration cost of $40$ steps for both segments. Additionally we use a planning horizon of $10$ for \algabbr and $30$ for PETSfD. However, because there is an intersection in the middle of the desired trajectory, \algabbr finds a shortcut to the goal state. Thus, the trajectory is divided into non-intersecting segments before \algabbr separately optimizes each one. At execution-time, the segments are stitched together and we find that \algabbr is robust enough to handle the uncertainty at the transition point. We hypothesize that this is because the dynamics and value function exhibit good generalization.

Results for both segments of the Figure 8 task are shown in Figures \ref{fig8-plots} and \ref{fig8-other-seg} below. In Figure \ref{fig8-plots}, we see that \algabbr quickly learns to smooth out demo trajectories while satisfying constraints, with a success rate of over $80 \%$ while baselines violate constraints on nearly every iteration and never complete the task, as shown in Figure \ref{fig8-plots}. Note that PETSfD almost always violates constraints, even though constraints are enforced exactly as in \algabbr. We hypothesize that since we need to give PETSfD a long planning horizon to make it possible to complete the task (since it has no value function), this makes it unlikely that a constraint satisfying trajectory is sampled with CEM. For the other segment of the Figure-8, \algabbr still quickly learns to smooth out demo trajectories while satisfying constraints, with a success rate of over $80 \%$ while baselines violate constraints on nearly every iteration and never complete the task, as shown in Figure \ref{fig8-other-seg}. 

\begin{figure*}
\centering
\begin{subfigure}[t]{.31\textwidth}
  \centering
  \includegraphics[width=\linewidth]{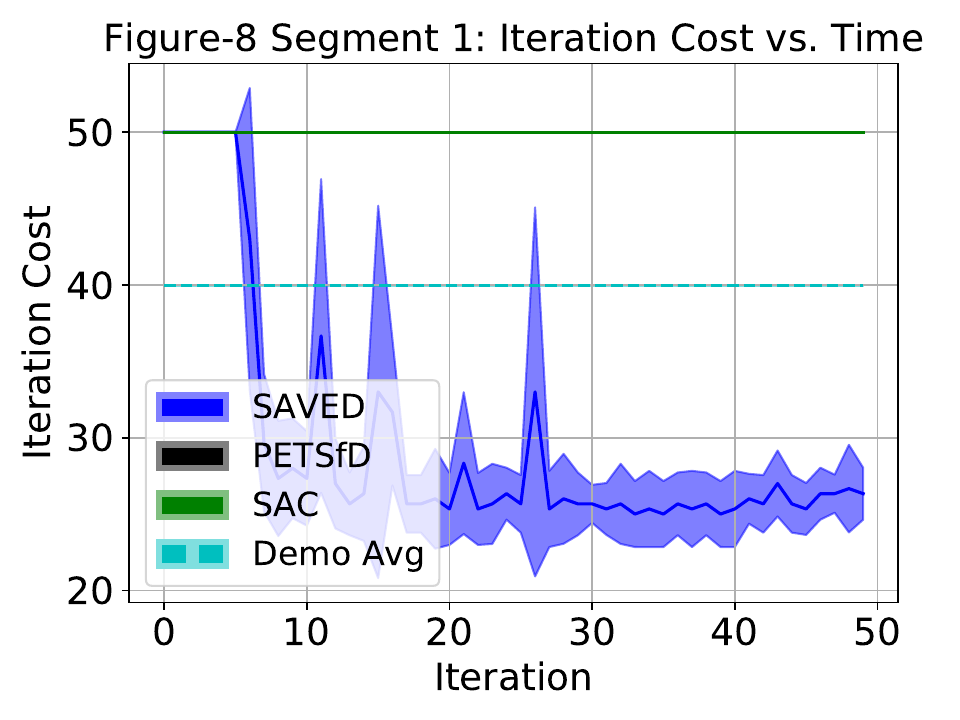}
  \label{fig8-seg1-returns}
\end{subfigure}
\hfill
\begin{subfigure}[t]{.31\textwidth}
  \centering
  \includegraphics[width=1.1\linewidth]{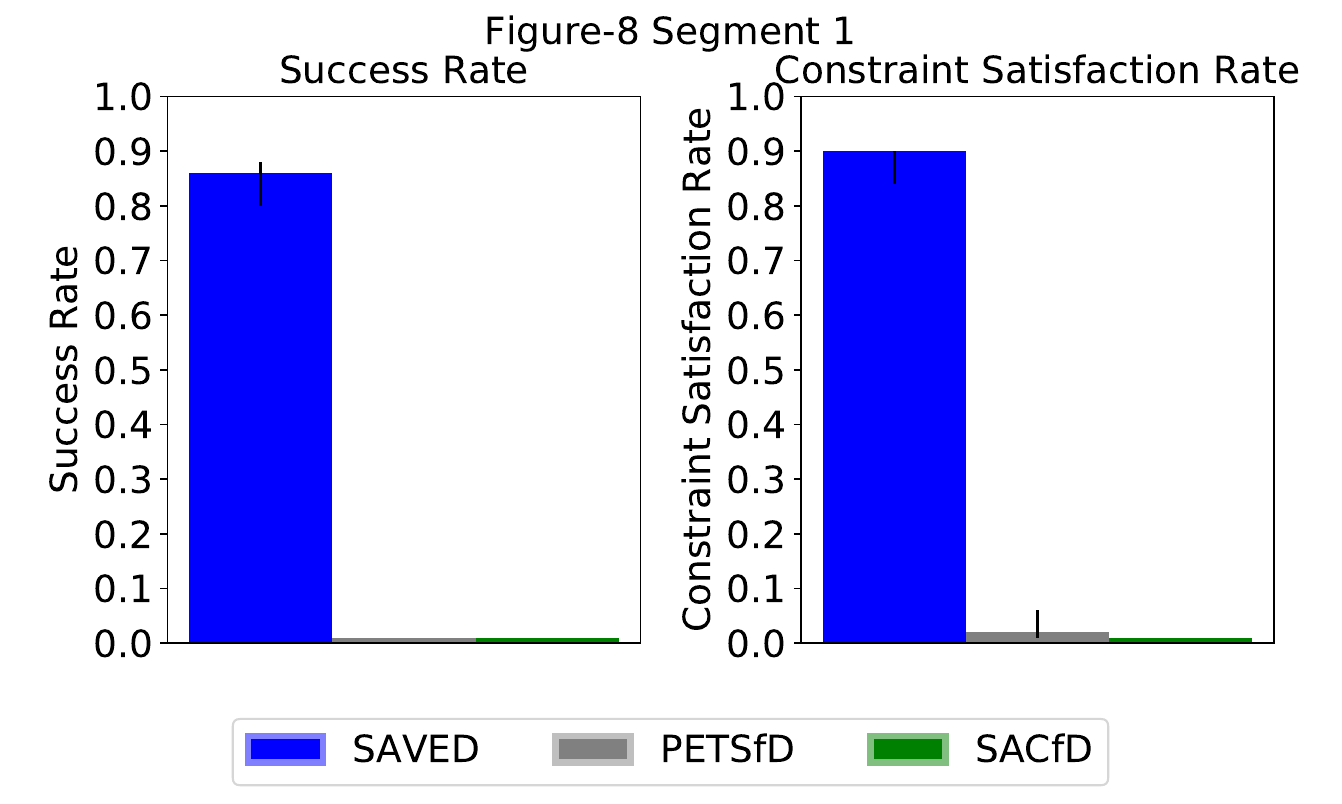}
  \label{fig8-seg1-stats}
\end{subfigure}%
\hfill
\begin{subfigure}[t]{.31\textwidth}
  \centering
  \includegraphics[width=\linewidth]{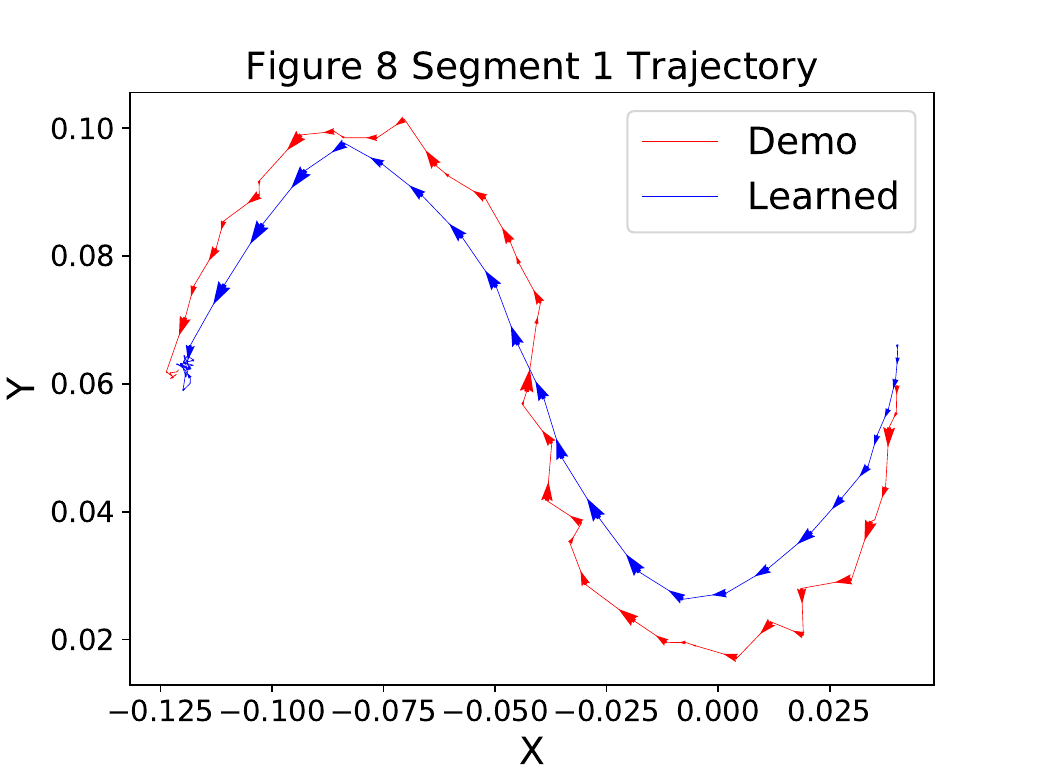}
  \label{fig8-seg1-trajs}
\end{subfigure}
\caption{\textbf{Figure-8: }\textbf{Training Performance: }After just $10$ iterations, \algabbr consistently succeeds and converges to an iteration cost of $26$, faster than demos which took an average of $40$ steps. Neither baseline ever completes the task in the first $50$ iterations; \textbf{Trajectories: }Demo trajectories satisfy constraints, but are noisy and inefficient. \algabbr learns to speed up with only occasional constraint violations and stabilizes in the goal set.}
\label{fig8-plots}
\end{figure*}

\begin{figure}
\centering
\begin{subfigure}[t]{.31\textwidth}
  \centering
  \includegraphics[width=0.8\linewidth]{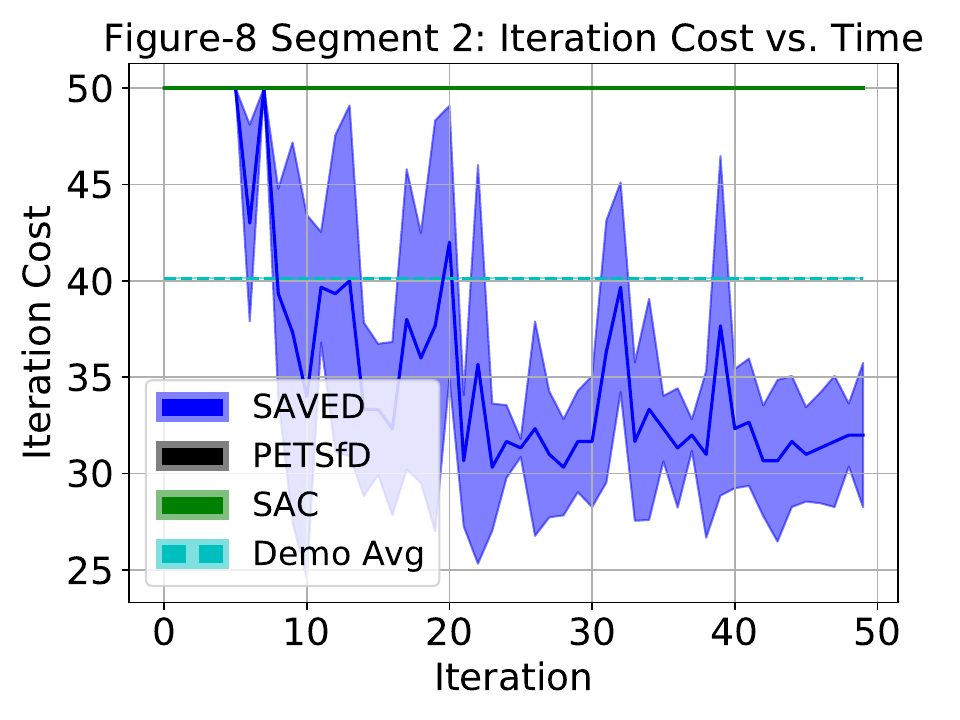}
  \label{fig8-seg2-returns}
\end{subfigure}
\hfill
\begin{subfigure}[t]{.31\textwidth}
  \centering
  \includegraphics[width=\linewidth]{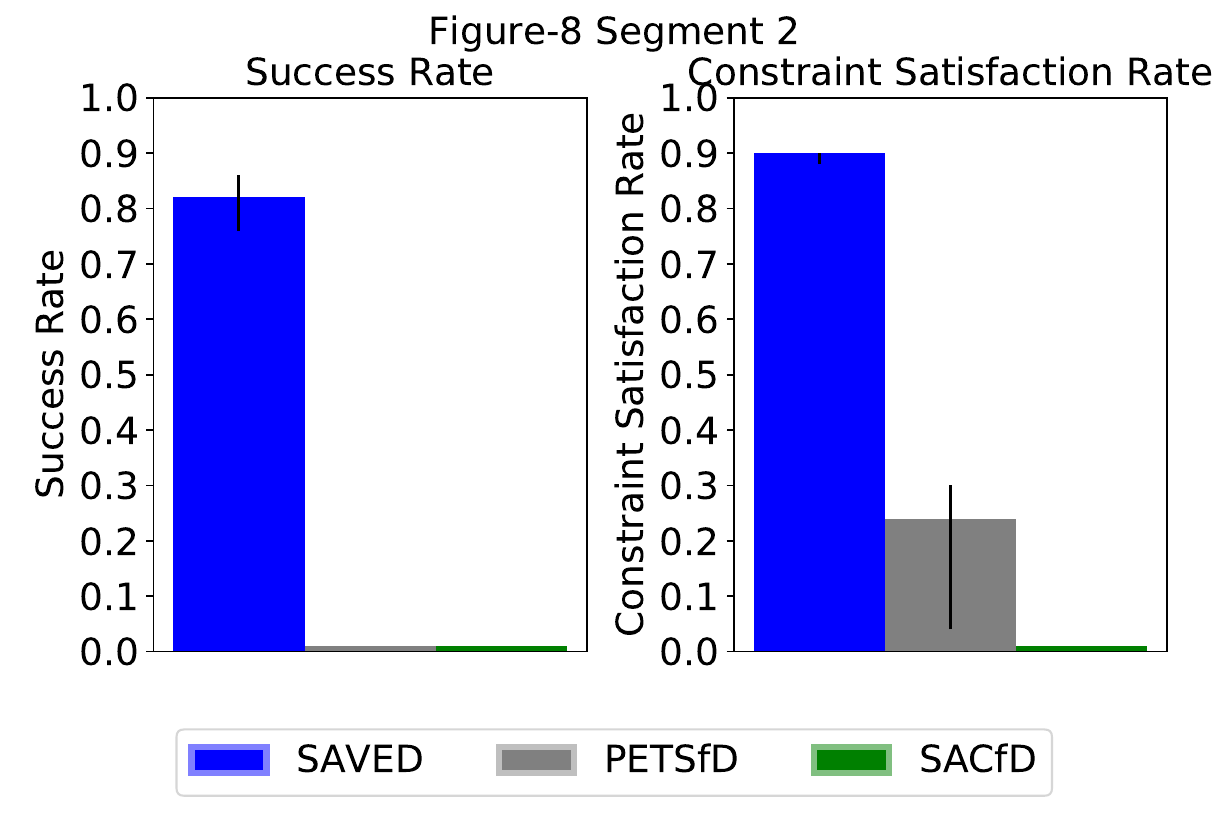}
  \label{fig8-seg2-stats}
\end{subfigure}%
\hfill
\begin{subfigure}[t]{.31\textwidth}
  \centering
  \includegraphics[width=0.8\linewidth]{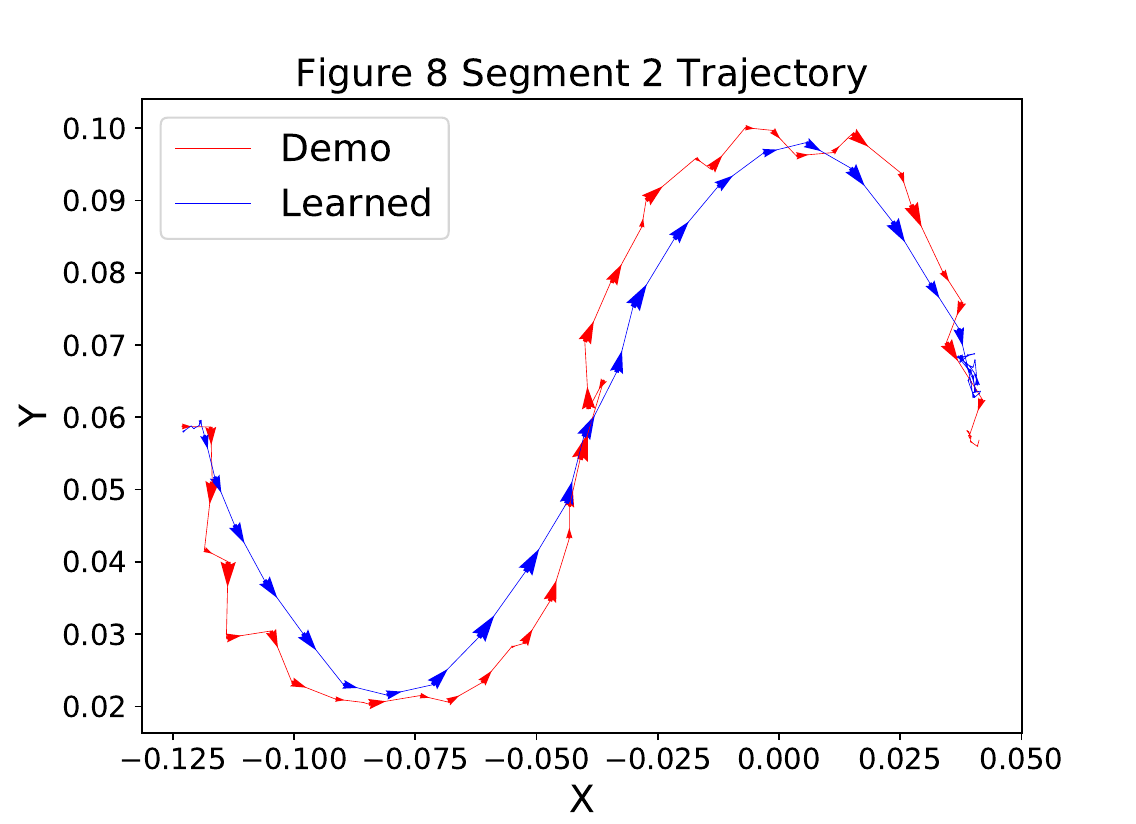}
  \label{fig8-seg2-trajs}
\end{subfigure}
\caption{\textbf{Figure-8: } \textbf{Training Performance: }After $10$ iterations, the agent consistently completes the task and converges to an iteration cost of around $32$, faster than demos which took an average of $40$ steps. Neither baseline ever completed the task in the first $50$ iterations; \textbf{Trajectories: }Demo trajectories are constraint-satisfying, but noisy and inefficient. \algabbr quickly learns to speed up demos with only occasional constraint violations and successfully stabilizes in the goal set. Note that due to the difficulty of the tube constraint, it is hard to avoid occasional constraint violations during learning, which are reflected by spikes in the iteration cost.}
\label{fig8-other-seg}
\end{figure}

\begin{figure}
  \centering
  \includegraphics[width=0.35\linewidth]{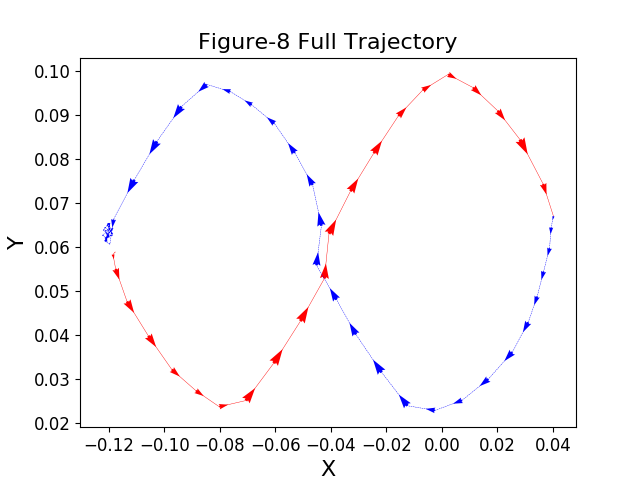}
  \caption{\textbf{Full Figure-8 trajectory:} We show the full figure-8 trajectory, obtained by evaluating learned policies for the first and second figure-8 segments in succession. Even when segmenting the task, the agent can smoothly interpolate between the segments, successfully navigating the uncertainty in the transition at execution-time and stabilizing in the goal set.}
  \label{fig8-full-learned-traj}
\end{figure}

In Figure \ref{fig8-full-learned-traj}, we show the full trajectory for the Figure-8 task when both segments are combined at execution-time. This is done by rolling out the policy for the first segment, and then starting the policy for the second segment as soon as the policy for the first segment reaches the goal set. We see that even given uncertainty in the dynamics and end state for the first policy (it could end anywhere in a 1 cm ball around the goal position), \algabbr is able to smoothly navigate these issues and interpolate between the two segments at execution-time to successfully stabilize at the goal at the end of the second segment. Each segment of the trajectory is shown in a different color for clarity. We suspect that \algabbr's ability to handle this transition is reflective of good generalization of the learned dynamics and value functions.

\subsection{Knot-Tying}
\begin{figure}
\centering
\begin{subfigure}[t]{.47\textwidth}
  \centering
  \includegraphics[width=0.8\linewidth]{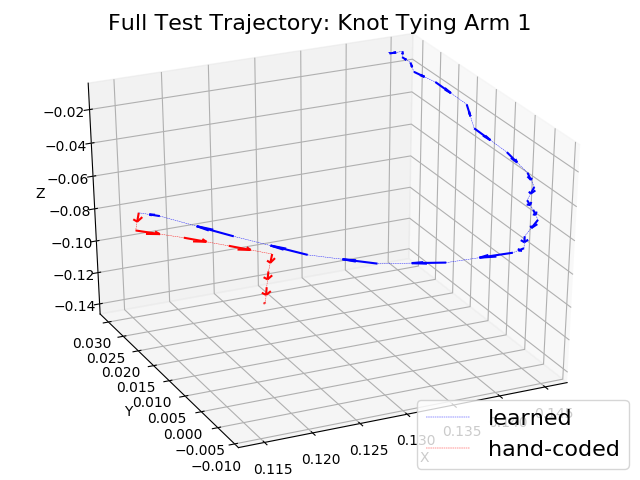}
  \label{knot-tying-arm1-full-traj}
\end{subfigure}
\hfill
\begin{subfigure}[t]{.47\textwidth}
  \centering
  \includegraphics[width=0.8\linewidth]{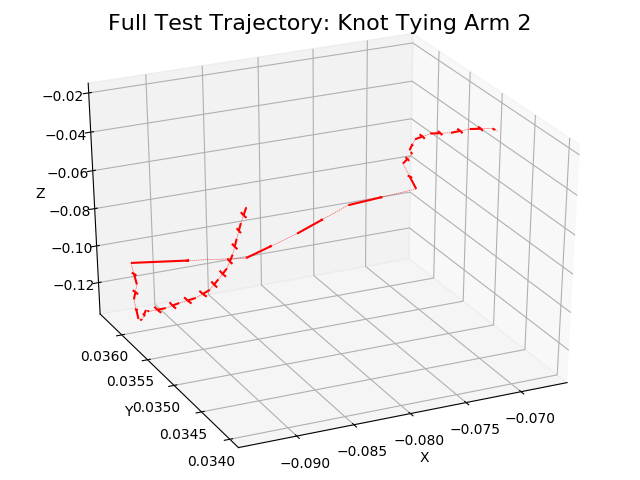}
  \label{knot-tying-arm2-full-traj}
\end{subfigure}%
\caption{\textbf{Knot-Tying Full Trajectories: } \textbf{(a) Arm 1 trajectory: }We see that the learned part of the arm 1 trajectory is significantly smoothed compared to the demonstrations at execution-time as well, consistent with the training results. Then, in the hand-coded portion of the trajectory, arm 1 is simply moved down towards the phantom along with arm 2, which grasps the thread and pulls it up; \textbf{(b) Arm 2 trajectory: }This trajectory is hand-coded to move down towards the phantom after arm 1 has fully wrapped the thread around it, grasp the thread, and pull it up.}
\label{full-knot-tying}
\end{figure}

In Figure \ref{full-knot-tying}, we show the full trajectory for both arms for the surgical knot-tying task. We see that the learned policy for arm 1 smoothly navigates around arm 2, after which arm 1 is manually moved down along with arm 2, which grasps the thread and pulls it up through the resulting loop in the thread, completing the knot.

\section{Ablations}
\subsection{\algabbr}

We investigate the impact of kernel width $\alpha$, chance constraint parameter $\beta$, and the number of demonstrator trajectories used on navigation task 2. Results are shown in Figure \ref{ablations}. We see that \algabbr is able to complete the task well even with just 20 demonstrations, but is more consistent with more demonstrations. We also notice that \algabbr is relatively sensitive to the setting of kernel width $\alpha$. When $\alpha$ is set too low, we see that \algabbr is overly conservative, and thus can barely explore at all. This makes it difficult to discover regions near the goal set early on and leads to significant model mismatch, resulting in poor task performance. Setting $\alpha$ too low can also make it difficult for \algabbr to plan to regions with high density further along the task, resulting in \algabbr failing to make progress. On the other extreme, making $\alpha$ too large causes a lot of initial instability as the agent explores unsafe regions of the state space. Thus, $\alpha$ must be chosen such that \algabbr is able to sufficiently explore, but does not explore so aggressively that it starts visiting states from which it has low confidence in being able reach the goal set. Reducing $\beta$ allows the agent to take more risks, but this results in many more collisions. With $\beta = 0$, most rollouts end in collision or failure as expected. In the physical experiments, we find that allowing the agent to take some risk during exploration is useful due to the difficult tube constraints on the feasible state space. 

\begin{figure}[H]
\centering
\begin{subfigure}[t]{0.47\textwidth}
  \centering
    \includegraphics[width=\linewidth]{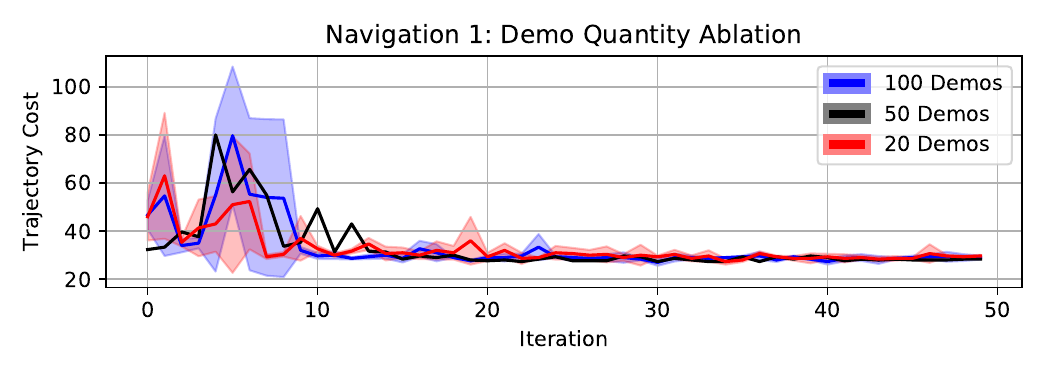}
\end{subfigure}%
\hfill
\begin{subfigure}[t]{0.47\textwidth}
  \centering
    \includegraphics[width=\linewidth]{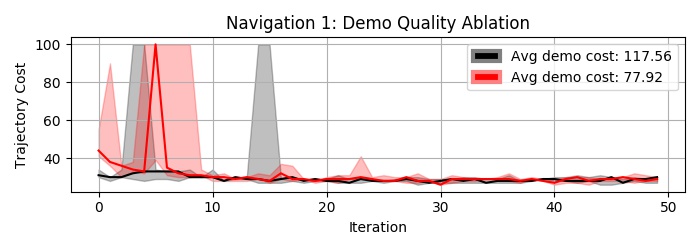}
\end{subfigure}
\caption{\textbf{\algabbr Ablations on Navigation Task 1: Number of Demonstrations:} SAVED is able to consistently complete the task with just both demo qualities considered without significant performance decay. The $100$ demonstrations provided in this task have average trajectory cost of $117.56$ (black) and $77.82$ (red) and SAVED significantly outperforms both, converging in less than $10$ iterations in all runs to a policy with trajectory cost less than $30$. \textbf{Demonstration quality:} SAVED is able to consistently complete the task with just 20 demonstrations (red) after 15 iterations. The demonstrations provided in this task have average trajectory cost of $77.82$.}
\label{ablations1}
\end{figure}

\begin{figure}[H]
\centering
\begin{subfigure}[t]{0.47\textwidth}
  \centering
    \includegraphics[width=\linewidth]{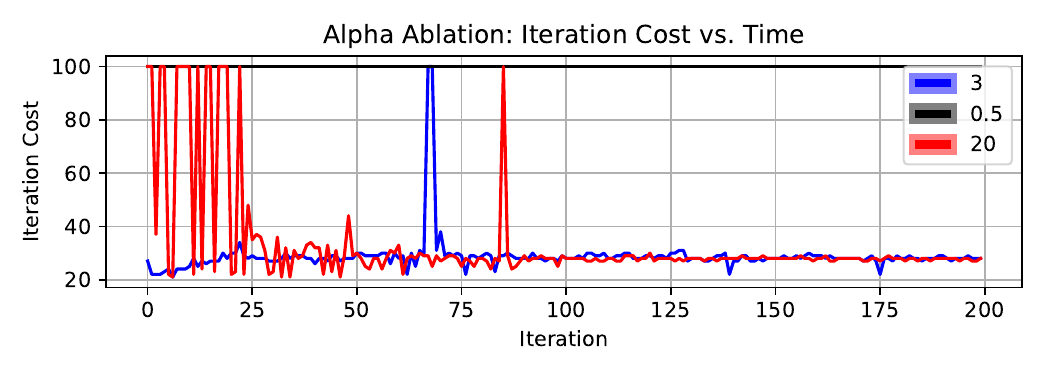}
\end{subfigure}%
\hfill
\begin{subfigure}[t]{0.47\textwidth}
  \centering
    \includegraphics[width=\linewidth]{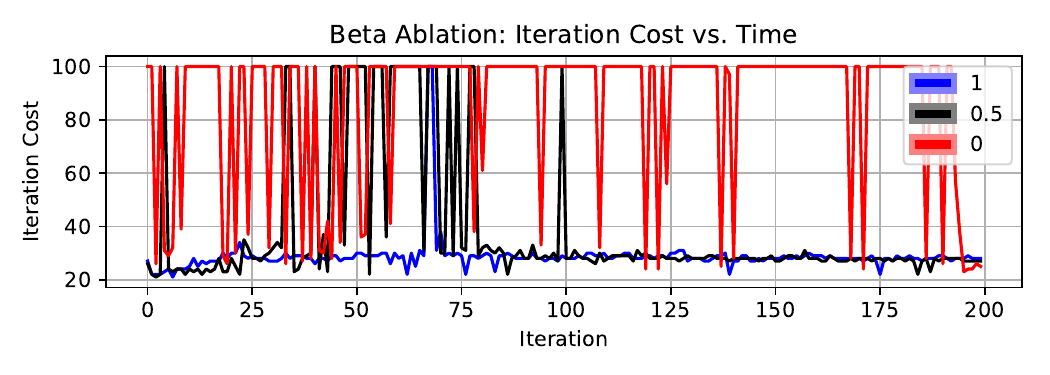}
\end{subfigure}
\vfill
\centering
\begin{subfigure}[t]{0.47\textwidth}
  \centering
    \includegraphics[width=\linewidth]{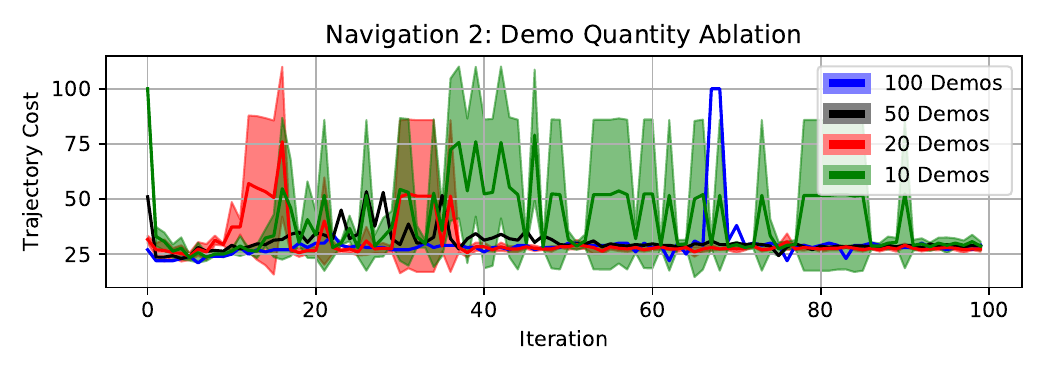}
\end{subfigure}
\begin{subfigure}[t]{0.47\textwidth}
\includegraphics[width=0.99\textwidth]{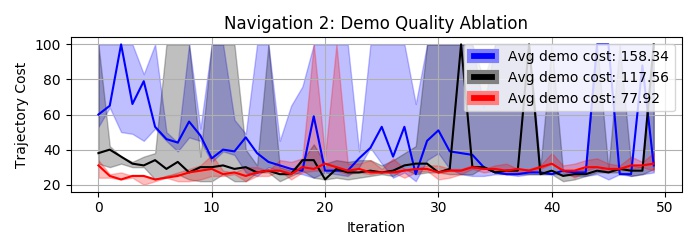}
\end{subfigure}
\caption{\textbf{\algabbr Ablations on Navigation Task 2:} \textbf{Kernel width $\alpha$: } We see that $\alpha$ must be chosen to be high enough such that \algabbr is able to explore enough to find the goal set, but not so high that \algabbr starts to explore unsafe regions of the state space; \textbf{Chance constraint parameter $\beta$:} Decreasing $\beta$ results in many more collisions with the obstacle. Ignoring the obstacle entirely results in the majority of trials ending in collision or failure. \textbf{Demonstration quantity: } In this experiment, we vary the number of demonstrations that SAVED is provided. We see that SAVED is able to complete the task with just 20 demonstrations (red), but more demonstrations result in increased stability during learning. Even with 10 demonstrations (green), SAVED is able to sometimes complete the task. The demonstrations provided in this task have average trajectory cost of $77.82$. \textbf{Demonstration quality:} SAVED efficiently learns a controller in all runs in all cases, the worst of which has demos that attain an iteration cost 5 times higher than the converged controller. We do occasionally observe some instability in the value function, which begins to display somewhat volatile behavior after initially finding a good controller. Constraints are never violated during learning in any of the runs.}
\label{ablations}
\end{figure}

Finally, we also ablate the quantity and quality of demonstrations used for navigation task 2 (Figure~\ref{ablations}), and find that SAVED is still able to consistently complete the task with just 20 demonstrations and is relatively robust to lower quality demonstrations, although this does result in some instability during training. We additionally ablate the quantity and quality of demonstrations for navigation task 1 in Figure~\ref{ablations1}. We note that again, SAVED is relatively robust to varying demonstration quality, achieving similar performance even for very slow demonstrations

\subsection{Model-Free}
\label{mf-ablation}

\begin{figure}[H]
  \centering
    \includegraphics[width=0.99\linewidth]{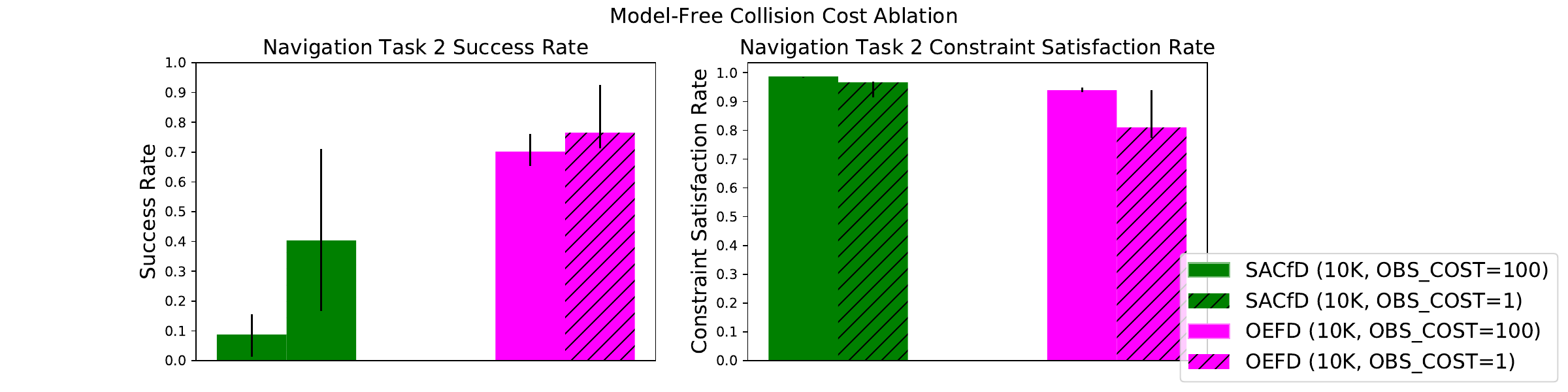}
    \caption{A high cost for constraint violations results in conservative behavior that learns to avoid the obstacle, but also makes it take longer to learn to perform the task. Setting the cost low results in riskier behavior that more often achieves task success.}
  \label{ablation-sac}
\end{figure}

\begin{figure}[H]
\centering
\begin{subfigure}[t]{0.47\textwidth}
  \centering
    \includegraphics[width=\linewidth]{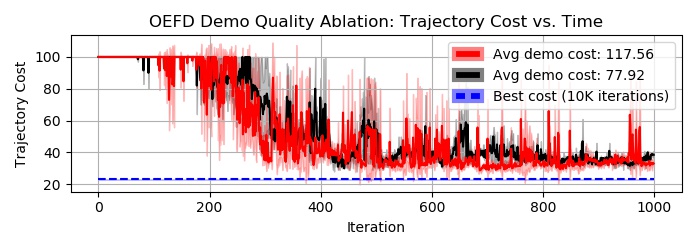}
\end{subfigure}%
\hfill
\begin{subfigure}[t]{0.47\textwidth}
  \centering
    \includegraphics[width=\linewidth]{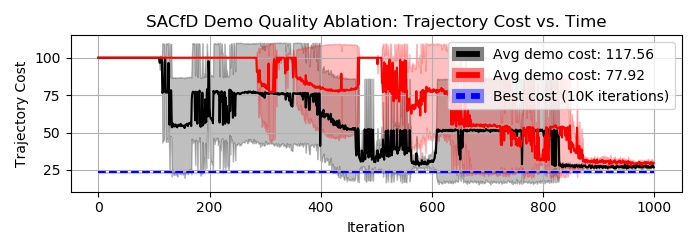}
\end{subfigure}
\caption{\textbf{\algabbr Model-Free RL Demo Quality Ablations on Navigation Task 1:} \textbf{OEFD:} We see that the baseline OEFD has similar performance across demonstration qualities. OEFD takes hundreds of iterations to start performing the task successfully, while SAVED converges less than 10 iterations; \textbf{SACfD:} We see that the baseline SACfD does slightly better with worse demonstrations. This could be due to the fact that more samples are placed in the agent's replay buffer with longer demonstrations. We note that both cases take hundreds of iterations to start completing the task, while SAVED starts to the complete the task almost immediately.}
\label{ablation-mf-quality}
\end{figure}

\begin{figure}[H]
\centering
\vspace{10pt}
\includegraphics[width=0.6\textwidth]{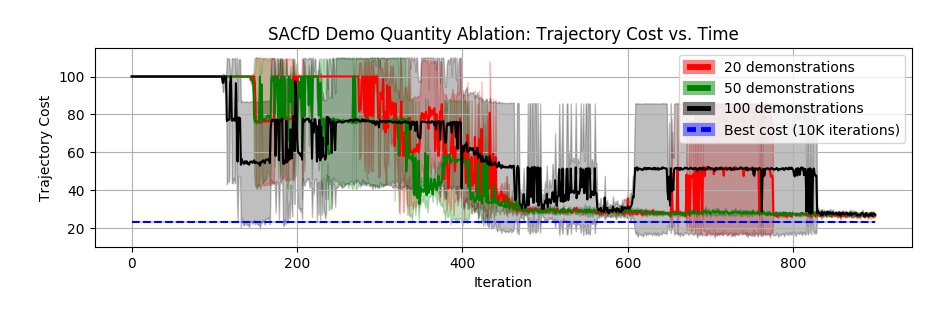}
\caption{\textbf{SACfD Demo Quantity Ablation on Navigation Task 1:} We study the effect of varying demonstration numbers on the model free RL baseline algorithm SACfD~\cite{SAC}. We see that the baseline SACfD has high variance across all demonstration quantities, and takes roughly similar time to converge in all settings with $50$ demonstrations (green) being the fastest. We also plot the best observed cost in 10,000 iterations across all runs (dashed blue) and note that unlike OEFD (Figure~\ref{ablation-mf-quality}), the SACfD runs all converge close to this value.}
\label{fig:sac_ablation_quantity}
\end{figure}

To convey information about constraints to model-free methods, we provide an additional cost for constraint violations. We ablate this parameter for navigation task 2 in Figure~\ref{ablation-sac}. We find that a high cost for constraint violations results in conservative behavior that learns to avoid the obstacle, but also takes much longer to achieve task success. Setting the cost low results in riskier behavior that succeeds more often. This trade-off is also present for model-based methods, as seen in the prior ablations. Additionally, we also ablate the demonstration quality for the model-free baselines, and find that increasing the iteration cost of the demonstrations by almost $50\%$ does not significantly change the learning curve of OEFD  (Figure~\ref{ablation-mf-quality}), and in both cases, OEFD takes much longer than SAVED to start performing the task successfully (Figure~\ref{ablations1}). We also perform the same study on the model-free RL baseline algorithm Soft Actor Critic from Demonstrations (SACfD)~\cite{SAC}. We observe that increasing demonstration length results in somewhat faster learning, and hypothesize this could be due to the replay buffer having more data to initially train from. We note that this method has high variance across the runs, and all runs took close to $900$ iterations to converge (Figure~\ref{ablation-mf-quality}) while SAVED converges in less than $10$ iterations (Figure~\ref{ablations1}). We also ablate demo quantity for SACfD on navigation task 1 in Figure~\ref{fig:sac_ablation_quantity} and find that although SACfD has a performance improvement with additional demonstrations, it takes a few hundred iterations to converge and more than a 100 iterations to even complete the task, while SAVED converges within 15 iterations (Figure~\ref{ablations1}).

